\documentclass[11pt]{article}

\usepackage[utf8]{inputenc} 
\usepackage[T1]{fontenc}    
\usepackage{hyperref}       
\usepackage{url}            
\usepackage{booktabs,natbib}       
\usepackage{amsfonts}       
\usepackage{nicefrac}       
\usepackage{microtype}      
\usepackage{pifont,xspace,paralist,times,amsthm}
\usepackage{titlesec}
\usepackage{amsmath,amssymb,times,algorithmic,fullpage}
\usepackage{latexsym,paralist,multirow,tablefootnote}
\usepackage{hyperref}
\hypersetup{colorlinks=true,linkcolor=blue,citecolor=blue}
\usepackage{graphicx}
\usepackage{amsfonts}
\usepackage{makecell}
\usepackage{caption}

\newtheorem{step}{Step}
\newtheorem{theorem}{Theorem}[section]
\newtheorem*{non-theorem}{Informal Theorem}
\newtheorem{lemma}[theorem]{Lemma}

\newtheorem{proposition}[theorem]{Proposition}

\newtheorem{remark}[theorem]{Remark}

\newtheorem{corollary}[theorem]{Corollary}
\newtheorem{definition}{Definition}

\def \y {\mathbf y}

\def \Lasso {\mathrm {Lasso}}

\def \z {\mathbf z}

\def \XXX {\mathcal{X}}

\def \YYY {\mathcal{Y}}
\def \NNN {\mathcal{N}}

\def \u {\mathbf u}
\def \N {\mathbb N}

\def \comp {{\mathrm{comp}}}

\def \supp {\mathrm{supp}}

\def \a {\mathbf a}

\def \compl {\mathrm{co}}
\def \e {\mathbf e}
\def \f {\mathbf f}

\def \g {\mathbf g}

\def \v {\mathbf v}
\def \x {\mathbf x}
\def \w {\mathbf w}

\def \M { M}
\def \R {\mathbb{R}}
\def \N {\mathbb{N}}

\def \NN {\mathcal{N}}

\DeclareMathOperator{\E}{\mathbb{E}}

\DeclareMathOperator{\diag}{diag}
\DeclareMathOperator{\Span}{span}
\DeclareMathOperator{\sr}{\mathbf{sr}}
\DeclareMathOperator{\tr}{tr}
\DeclareMathOperator{\rank}{rank}
\DeclareMathOperator{\conv}{conv}
\def \HS {\mathrm{F}}
\def \< {\left \langle}
\def \> {\right \rangle}
\def \etc {,\ldots,}
\newcommand{\norm}[1]{\left \| #1 \right \|}
\newcommand{\Prob}[2][]{\mathrm{Pr}_{#1} \left[ #2 \rule{0mm}{3mm}\right]}
\def \RE {\textsc{Re}}

\def \< {\left \langle}
\def \> {\right \rangle}

\begin{document}
\title{Restricted Eigenvalue from Stable Rank \\with Applications to Sparse Linear Regression}
 \author{Shiva Kasiviswanathan\thanks{Amazon AWS AI, Palo Alto, CA, USA. \href{mailto:kasivisw@gmail.com}{kasivisw@gmail.com}.} \and Mark Rudelson\thanks{University of Michigan, Ann Arbor, MI, USA. \href{mailto:rudelson@umich.edu }{rudelson@umich.edu }.  Partially supported by NSF grant, DMS-1464514.}}
\date{}
\maketitle
\vspace*{-6ex}
\begin{abstract}
High-dimensional settings, where the data dimension ($d$) far exceeds the number of observations ($n$), are common in many statistical and machine learning applications. Methods based on $\ell_1$-relaxation, such as Lasso, are very popular for sparse recovery in these settings. Restricted Eigenvalue (RE) condition is among the weakest, and hence the most general, condition in literature imposed on the Gram matrix that guarantees nice statistical properties for the Lasso estimator. It is natural to ask: what families of matrices satisfy the RE condition?  Following a line of work in this area~\citep{raskutti2010restricted,rudelson2013reconstruction,sivakumar2015beyond,oliveira2016lower,lecue2017sparse}, we construct a new broad ensemble of dependent random design matrices that have an explicit RE bound. Our construction starts with a fixed (deterministic) matrix $X \in \R^{n \times d}$ satisfying a simple stable rank condition, and we show that a matrix drawn from the distribution $X \Phi^\top \Phi$, where $\Phi \in \R^{m \times d}$ is a subgaussian random matrix, with high probability, satisfies the RE condition. This construction allows incorporating a fixed matrix that has an easily {\em verifiable} condition into the design process, and allows for generation of {\em compressed} design matrices that have a lower storage requirement than a standard design matrix. We give two applications of this construction to sparse linear regression problems, including one to a compressed sparse regression setting where the regression algorithm only has access to a compressed representation of a fixed design matrix $X$.
\end{abstract}

\section{Introduction} \label{sec:intro}
A high dimensional setting, where the number of features ($d$) is much larger than the number of observations ($n$) appears commonly in statistics and signal processing, for example, in regression, covariance selection on Gaussian graphical models, signal reconstruction, and sparse approximation. Consider a simple setting where we try to recover $\theta^\star$, given $(\M,\y)$, satisfying the following linear model: 
\begin{equation} \label{eqn:linear}
\y = \M \theta^\star + \w.
\end{equation} 
Here $\y \in \R^n$ is the vector of noisy observations, $\M \in \R^{n \times d}$ is the design matrix, and $\w \in \R^n$ is an unknown noise vector. In the setting of $d \gg n$, the model is {\em unidentifiable} and it is not meaningful to estimate $\theta^\star\in \R^d$.  However, many machine learning and statistical applications, exhibit special structure that can lead to an identifiable model. In particular, in many settings, the vector $\theta^\star$ is sparse. Given such a problem, the most direct approach would be to seek an exact sparse minimizer of the least-squares cost, $\|\y - \M \theta\|^2$, thereby obtaining an $\ell_0$-based estimator. However, since this problem is non-convex, a standard approach is to replace the $\ell_0$-constraint  with its $\ell_1$-norm which is the basis for methods such as Lasso~\citep{tibshirani1996regression} and Dantzig selector~\citep{candes2007dantzig}. There is now a well-developed theory of what conditions on the design matrix $\M$ are needed for these $\ell_1$-based relaxations to succeed. The general idea is that $\M$ needs to behave sufficiently nicely in a sense that it satisfies certain {\em incoherence} conditions. One popular notion of incoherence is {\em Restricted Isometry Property} (RIP) that states for all $k$-sparse sets $T \subset \{1,\dots,d\}$ ($|T|=k$), the matrix $\M$ restricted to the columns from $T$ acts as an almost isometry~\citep{candes2005decoding}. In the past decade, few variants of the RIP notion for exact and approximate recovery of $\theta^\star$, under the noiseless and noisy setting, have also been proposed (we refer to reader to the books by~\citep{eldar2012compressed,hastie2015statistical} for more details). 

For the Lasso and Dantzig selector,~\citep{bickel2009simultaneous} formulated the {\em restricted eigenvalue} (RE) condition and showed that it is the among the weakest,\!\footnote{In particular~\citep{bickel2009simultaneous} show that the RE condition is a relaxation of the RIP condition under suitable choices of parameters involved in both of them.} and hence the most general, condition imposed on the Gram matrix that guarantees meaningful recovery. Informally, the RE condition on a matrix $M$ involves lower bounds on $\| M \theta \|$ that hold uniformly over an appropriately defined subset of sparse vectors (see Definition~\ref{def:RE} for a formal statement). A natural question is then: for what ensembles of design matrices does the restricted eigenvalue condition hold (say, with high probability)? Standard constructions satisfying the RE condition are based on i.i.d.\ random matrices,  independent draws from a set of uncorrelated basis functions,  additive combinatorics, or coding-theoretic techniques
(see, e.g.,~\citep{mendelson2008uniform,adamczak2011restricted,rudelson2008sparse,bourgain2011explicit,cheraghchi2011coding} and references therein). While these constructions are well-suited for certain compressive sensing tasks, where we have control over the design matrix, it may not be appropriate for statistical inference problems such as sparse linear regression, where the design matrix is not under control of the ``experimenter''. For example, it is common that the different columns (covariates) of the design matrix are correlated with one other, and in practice $\ell_1$-norm methods such as Lasso seem to perform well even in these settings.  This has motivated recent work in understanding RE properties for a more realistic class of random design matrices~\citep{raskutti2010restricted,rudelson2013reconstruction,sivakumar2015beyond,oliveira2016lower,lecue2017sparse}. Our paper continues this line of work.

We start with this simple question: can we incorporate a fixed (deterministic) matrix while constructing a family of matrices satisfying the RE condition? In this paper, we answer this question in affirmative by presenting a construction that starts with any deterministic matrix $X \in \R^{n \times d}$,  satisfying a very mild easy to check condition, and generates a distribution of matrices centered at $X$, such that a matrix drawn from this distribution with high probability satisfies the RE condition. More formally, we show that given $X$, a matrix drawn from the distribution $X \Phi^\top \Phi$, where $\Phi \in \R^{m \times d}$ is a subgaussian random matrix, satisfies the RE condition with high probability.\!\footnote{We overload notation and use $X \Phi^\top \Phi$ to represent both a random sample and its distribution.} All we need is that the {\em stable rank} of $X$ is not ``too small''.  Stable rank of a matrix $X$ (denoted by $\sr(X)$), defined as the squared ratio of Frobenius and spectral norms of $X$, is a commonly used robust surrogate to usual matrix rank in linear algebra.  We start with an informal statement of our main result which shows, that under some mild conditions on $X$, with high probability $X \Phi^\top \Phi$ satisfies the restricted eigenvalue property with a parameter value of $\| X \|_F^2 /n m k$. 
\begin{non-theorem}[See Theorem~\ref{thm:RE}, Corollary~\ref{cor:RE}]
Let $X$ be a fixed $n \times d$ matrix with stable rank greater than $m$. Let $\Psi$ be an $m \times d$ subgaussian matrix with i.i.d.\ entries and let $\Phi = \Psi/\sqrt{m}$, then for any $k$ such that $m \gtrsim k^2$, with high probability,
\begin{equation} \label{eqn:REthm}
\inf_{S \subset [d], |S| =  k, \theta \in  \mathbb{C}(S)}  \frac{ \| X \Phi^\top \Phi \theta \|^2}{n \| \theta \|^2 } \geq \frac{ \| X \|_F^2 }{n m k},
\end{equation}
where $\mathbb{C}(S)$ is the set of vectors $\theta \in \R^d$ that satisfy the cone constraint, $\mathbb{C}(S) = \{ \theta \,:\,  \| \theta_{S^\compl} \|_1 \leq 3\| \theta_{S} \|_1 \}$ and $\theta_{S}, \theta_{S^\compl}$ represents the subvector of $\theta$ confined to coordinates $S$ and $\{1,\dots,d\} \setminus S$.
\end{non-theorem}
The stable rank is independent of the coordinate system, unlike RE which is tied to a concrete coordinate structure. The randomness of $\Phi$ makes the required condition on $X$ coordinate independent. The above proof is challenging because applying standard concentration tools directly do not give strong enough probability estimates on this quantity for a fixed $\theta$ to successfully apply an $\varepsilon$-net argument. To overcome this problem, we develop an orthogonal projection idea that allows us to decouple dependencies and reduce the problem to a state that is amenable to an application of an $\varepsilon$-net argument. Throughout the proof, we rely on the Hanson-Wright inequality and several of its consequences. 

\smallskip
\noindent\textbf{Some Key Features of this Construction.} We now note some interesting features of the family of random matrices generated by our construction. Firstly, observe that the entries in a matrix $Z = X \Phi^\top \Phi$ are highly correlated with $\E[Z] = X$.  Given any matrix $X$, its stable rank can be computed easily. This is an important advantage while designing RE matrices, as this makes the construction process {\em verifiable}, i.e., with high probability we can generate a matrix that satisfies an explicit restricted eigenvalue parameter bound.  Note that in general, checking whether a matrix satisfies the RE condition is a NP-hard problem~\citep{dobriban2016regularity}. To date, the main routes for constructing design matrices with an  explicit restricted eigenvalue bound have been via taking i.i.d.\ random ensembles (under different moment or tail assumptions) or  constructions through coding-theoretic techniques (such as expander codes~\citep{de2014optimal}), both of which generate family of matrices whose assumptions are not always reasonable for machine learning applications.  To the best of our knowledge, this is first construction of a very broad family of (correlated) random matrices that starts with an easy to check condition on the deterministic core.  Previous constructions of other such broad family of correlated random designs, such as~\citep{raskutti2010restricted,rudelson2013reconstruction}, require the deterministic matrix to also satisfy some suitable RE condition (more discussion in Section~\ref{sec:related}), thus running into the above mentioned verifiability issues. 

An additional salient feature is that the matrix $Z$ can be stored using only $O(m(n+d)) = O(md)$ words of memory as the factorization pair $(X \Phi^\top, \Phi)$. This means that compared to a standard $n \times d$ design matrix which needs $O(nd)$ words of memory (with $n$ generally being much greater than $m$), the design matrices coming out of this construction have a ``compressed'' representation. This property is useful when working with large design matrices in presence of memory constraints. 

\begin{figure*}  
\begin{center}
\hspace*{4ex}
\includegraphics[scale=0.4]{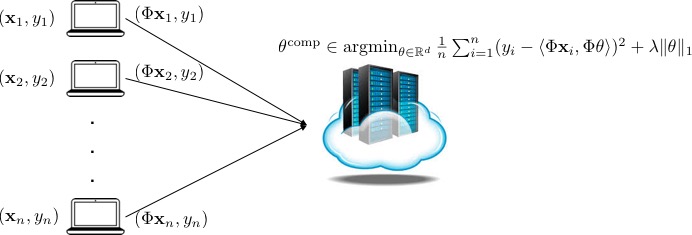}
\vspace*{-2ex}
\caption{A distributed data setting, where $n$ devices generating $(\x_1,y_1),\dots,(\x_n,y_n)$ are sending a compressed representation $(\Phi \x_1,y_1), \dots,(\Phi \x_n,y_n)$ to the cloud server, which then computes the  regression parameter.} 
\label{fig:skconv}
\end{center}
\vspace*{-5ex}
\end{figure*}

\smallskip
\noindent\textbf{Applications to Sparse Linear Regression.} We will give two applications of this result in sparse linear regression.  Consider the linear regression model in~\eqref{eqn:linear}. A popular approach for solving a (traditional) sparse linear regression problem is the Lasso technique of $\ell_1$-penalized regression. Lasso minimizes the usual mean squared error loss penalized with (a multiple of) the $\ell_1$-norm of $\theta$. The consistency properties of the Lasso estimator under various measurements of performance (such as prediction error, parameter error, support recovery) are now well-understood, see e.g.,~\citep{bickel2009simultaneous,wainwright2009sharp}. We consider the following Lasso problem, defined on the pair $(Z,\y)$, where $Z = X \Phi^\top \Phi$.
\begin{equation*} 
\theta^\comp \in \mbox{argmin}_{\theta \in \R^d}\, \frac{1}{n} \| \y - Z \theta \|^2 +  \lambda \| \theta \|_1 =  \mbox{argmin}_{\theta \in \R^d}\, \frac{1}{n}  (y_i - \langle \Phi \x_i, \Phi \theta \rangle)^2 +  \lambda \| \theta \|_1.
\end{equation*}
For brevity, in the following discussion, we make some simplifying assumptions and omit dependence on all but key variables. The $i$th row in $X$ ($\x_i^\top$) represent the covariates for the $i$th observation, and $\y = (y_1,\dots,y_n)$. The results stated below all are high probability bounds.
%
\begin{list}{{\bf (\arabic{enumi})}}{\usecounter{enumi}
\setlength{\leftmargin}{8pt}
\setlength{\listparindent}{3pt}
\setlength{\parsep}{3pt}}
\item \textbf{Parameter bound with design matrix $Z$.} Our first application is for the linear model $\y = Z \theta^\star + \w$. In this setting, we have a random design matrix. Here, the RE result on $Z$ leads to a parameter error bound: $\| \theta^\comp - \theta^\star \| = O(\sqrt{m} k^{3/2}/\| X \|_F)$, assuming $\w$ is an i.i.d.\ subgaussian noise vector (see Proposition~\ref{prop:app1}). While this result follows from a simple instantiation of the standard Lasso analysis framework, the result shows that there exists a new broad class of random design matrices for which Lasso succeeds in getting a consistent estimate of $\theta^\star$ (when $\| X \|_F = \omega(\sqrt{m} k^{3/2})$). A similar analysis can also be carried for the Dantzig selector based on the results of~\citep{bickel2009simultaneous} (omitted here).

\item \textbf{Sparse linear regression with compressed features.} Our second application is a variant of sparse linear regression. We start with a linear model $\y = X \theta^\star + \w$, where $X$ is a fixed matrix and $\w$ is an i.i.d.\ subgaussian noise vector (so in this case, we have fixed design $X$). However, we assume that the regression algorithm has access to only $(\Phi \x_1,y_1), \dots,(\Phi \x_n,y_n)$, which is the compressed representation of the original covariate-response pairs $(\x_1,y_1),\dots,(\x_n,y_n)$.\footnote{Note that given $\Phi \x_i$ it is not possible to accurately infer $\x_i$ without some strong (sparsity-like) assumptions on $\x_i$. More discussion on this is provided in Section~\ref{sec:compsparse}.}  Random projections are a class of extremely popular technique for dimensionality reduction (compression), where the original high-dimensional data is projected onto a lower-dimensional subspace using some appropriately chosen random matrix $\Phi$. A motivating scenario for this setting is as follows (see the illustration in Figure~\ref{fig:skconv}). Consider a distributed data setting, where $n$ devices each generating its own covariate-response pair is communicating to a central server (cloud).  If $d$ is large then communicating $\x_i \in\R^d$ is communication expensive. A natural scheme here is that the server chooses and announces a single random projection matrix $\Phi$, and every input point $\x_i$ can be compressed and sent as $\Phi\x_i$ to the server.  Such a scheme can be applied {\em locally} (i.e., on each $\x_i$ independent of the other), and reduces the overall communication by a factor of $d/m$.\!\footnote{We ignore the cost of communicating $\Phi$ to devices, which can be achieved using various techniques such as {\em one-to-all broadcasting}. In a practical implementation, $\Phi $ will be generated by a pseudorandom generator initialized by some seed, so by just communicating the seed we can regenerate $\Phi$ at each device. Also, with some small degradation in the parameters, the same $\Phi$ can be used in a situation where we have to repeatedly solve different sparse linear regression problem instances.} Now the goal of the server is to solve the regression problem for the original linear model ($\y = X \theta^\star + \w$) but from the available compressed representation. Firstly, since the $\x_i$'s are unavailable, it is {\em a priori} unclear how sparse linear regression performs in this setting. Secondly,  just with a stable rank condition on $X$ a parameter error bound on $\theta^\star$, that requires a stronger RE like assumption~\citep{raskutti2011minimax}, is ruled out. In this fixed design setting, we investigate (in-sample) prediction error bounds, and show that $\theta^\comp$ (which can be estimated from the compressed representation) satisfies $\| X \theta^\comp - X \theta^\star \|^2/n = O(\|X\|_F^2 k^{3/2}/nm)$ (see Proposition~\ref{prop:main}). In this case, our use of the RE slightly differs from the standard use of RE in Lasso analysis. We first bound $\| Z \theta^\comp - Z \theta^\star \|$ using the Lasso analysis framework, and then use the RE bound on $Z$ to relate that to a bound on $\| X \theta^\comp - X \theta^\star \|$.
\end{list}

\subsection{Related Work} \label{sec:related}
\noindent\textbf{Restricted Eigenvalue Bound.} Matrices that satisfy the restricted isometry (or restricted eigenvalue) property have many interesting applications in high-dimensional statistics and compressed sensing. However, there is no known way to generate them deterministically for a large range of sparsity levels $k$ (some of best constructions here include~\citep{bourgain2011explicit,bandeira2017conditional,bandeira2016derandomizing}), and additionally as discussed above these constructions lead to family of matrices that are not always reasonable for applications such as linear regression.  Interestingly though, it is easy to generate large matrices satisfying the RIP property (and therefore RE) for a wide range of sparsity levels through i.i.d.\ random design.  In statistics and machine learning, one common assumption is that the design matrix is generated randomly by a mechanism which is not under control of the experimenter, and these matrices generally have dependent entries. One may enquire whether such random ensembles will typically satisfy restricted eigenvalue properties. This problem was first addressed for Gaussian ensembles by~\citep{raskutti2010restricted} and then for subgaussian and bounded-coordinate ensembles by~\citep{rudelson2013reconstruction}. In particular, these results have established RE bounds for random matrices with subgaussian rows and non-trivial covariance structure, as well as random matrices with independent rows and uniformly bounded entries. Recent papers~\citep{sivakumar2015beyond,oliveira2016lower,lecue2017sparse} have developed variants of these bounds under different moment or tail assumptions. The closest relation to our work is the result by~\citep{rudelson2013reconstruction}, who showed that for a deterministic matrix $X$ satisfying the RE condition, the matrix $\Phi X$ satisfies the RE condition too (with a weaker RE parameter), where the rows of $\Phi$ are {\em isotropic} random vectors. Note that, unlike this result, we have a simple polynomial time checkable stable rank condition on our deterministic matrix $X$.

\noindent\textbf{Applications to Sparse Linear Regression.}  Lasso, is the most widely studied scheme for sparse linear regression. There has been a large and rapidly growing body of literature for Lasso and its variants which include theoretical explorations of its behavior and computationally efficient procedures for solving it. We refer the reader to the recent book by~\citep{hastie2015statistical} for a detailed survey about developments here. For applications of our RE bound to sparse linear regression, we draw on this rich literature studying theoretical properties of Lasso.

\citep{zhou2009compressed} considered sparse linear regression in a setting where the covariate matrix $X$ is pre-multiplied by a Gaussian random projection matrix to generate a reduced set of new datapoints in $d$-dimensions. They provide a convergence analysis of the Lasso estimator built from this compressed dataset. This setting is however different from ours, as we consider reducing the dimensionality of each covariate vector. In a high-dimensional setting, with $d \gg n$, reducing the dimensionality seems intuitively the more desirable way of achieving compression.

A recent area of research is that of distributed sparse linear regression where the dataset is assumed to the distributed across multiple machines. \citep{lee2015communication} showed that if the data is ``not too'' distributed, and for the random design case, average of individual Lasso estimators properly debiased converges to $\theta^\star$ at almost the same rate as the centralized Lasso estimator. We are not aware of a direct connection between this work and our setting.

\subsection{Preliminaries}
\noindent\textbf{Notation.} We denote $[n]=\{1,\ldots,n\}$. For a set $S \subseteq [d]$, $S^\compl$ denotes its complement set. Vectors are in column-wise fashion, denoted by boldface letters. For a vector $\v$, $\v^\top$ denotes its transpose, $\|\v\|_p$ it's $\ell_p$-norm, and $\supp(\v)$ its support. We use $\e_j \in \R^d$ to denote the standard basis vector with $j$th entry set to $1$. For a matrix $M$, $\|M\|$ denotes its spectral norm which equals its largest singular value, and $\| M \|_\HS$ its Frobenius norm. $\mathbb{I}_d$ represents the $d \times d$ identity matrix. For a vector $\x$ and set of indices $S$, let $\x_S$ be the vector formed by the entries in $\x$ whose indices are in $S$, and similarly, $X_S$ is the matrix formed by columns of $X$ whose indices are in $S$. The $d$-dimensional unit ball in $\ell_p$-norm centered at origin is denoted by $B_p^d$. The Euclidean sphere in $\R^d$ centered at origin is denoted by $\mathbb{S}^{d-1}$. We call a vector $\a \in \R^d$, {\em $k$-sparse}, if it has at most $k$ non-zero entries. Denote by $\Sigma_k$ the set of all vectors $\a \in B_2^d$ with support size at most $k$: $\Sigma_k=\{ \a \in B_2^d \,: \, | \supp(\a) | \leq k\}$. 

Throughout this paper, we assume covariate-response pairs come from some domain $\XXX \times \YYY$ where $\XXX \subset \R^d$ and  $\YYY \subset \R$. In Appendix~\ref{app:addl}, we also review a few additional concepts related to  sparse linear regression, $\varepsilon$-nets, and subgaussian random variables.

\smallskip
\noindent\textbf{RE and Sparse Linear Regression.} 
In the following, we consider the linear model: $\y = M \theta^\star + \w$. For a set $S \subset [d]$, let us define a cone set $\mathbb{C}(S)$ as:
$$\mathbb{C}(S) = \{ \theta \in \R^d \,: \, \| \theta_{S^\compl} \|_1 \leq 3\| \theta_{S} \|_1 \}.$$

Restricted eigenvalue is a mild condition on the covariate matrix that is sufficient for estimating $\theta^\star$ in a noisy linear model setup.\!\footnote{Given that we observe only a noisy version of the product $M \theta^\star$, it is then difficult to distinguish $\theta^\star$ from other sparse vectors. Thus, it is natural to impose an RE condition if the goal is to produce an estimate $\tilde{\theta}$ such that $\| \theta^\star-\tilde{\theta}\|$ is small.}
\begin{definition} [Restricted Eigenvalue~\citep{bickel2009simultaneous}] \label{def:RE}
A matrix $M \in \R^{n \times d}$ satisfies the restricted eigenvalue (RE) condition with parameter $\xi$ if,
$$ \inf_{S \subset [d], |S| =  k, \theta \in  \mathbb{C}(S)} \frac{\| M \theta \|^2}{ n} \geq  \xi \| \theta \|^2.$$
\end{definition}
Restricted eigenvalue is in fact a special case of a general property of loss functions, known as the {\em restricted strong convexity}, which imposes a type of strong convexity condition for some subset of vectors~\citep{negahban2012unified}. We now state a well-known result in sparse linear regression that provides a bound on the Lasso error, based on the linear observation model $\y = M\theta^\star + \w$. \begin{theorem}[\citep{bickel2009simultaneous}] \label{thm:lassoanalysis}
Let $\y = M\theta^\star + \w$ for a noise vector $\w \in \R^n$ and $\theta^\star$ is $k$-sparse.  Let $\lambda \geq 2\| M^\top \w \|_\infty/n$. Suppose $M$ satisfies the restricted eigenvalue condition with parameter $\xi > 0$,  then any optimal minimizer, $\tilde{\theta} \in \mbox{argmin}_{\theta \in \R^d}\, \frac{1}{n} \| \y - M \theta \|^2 + \lambda \| \theta \|_1$,  satisfies: $\| \tilde{\theta} - \theta^\star \| \leq 3 \sqrt{k} \lambda/\xi$.
\end{theorem}

\begin{remark} \label{rem:assumptions} [A Note on Assumptions] While the above RE condition is common for analyzing the $\ell_2$-error of the Lasso estimator, {\em stronger} conditions are used for achieving the {\em stronger} guarantee of {\em consistent support selection }~\citep{wainwright2009sharp}. These include  {\em mutual incoherence} and {\em minimum eigenvalue} conditions on $M$, and {\em minimum signal value} condition on $\theta^\star$. These conditions are known to be highly restrictive~\citep{tibshiranisparsity} and are not studied in this paper.
\end{remark}

\section{Restricted Eigenvalue from Stable Rank} \label{sec:REfromSR}
The main result of this paper is to show that the RE condition holds with high probability for systems of random design matrices of a general nature. In particular, we consider design matrices of the form: $Z = X \Phi^\top \Phi$, where $X$ is a fixed matrix and $\Phi \in \R^{m \times d}$ is a subgaussian random matrix. Note that the entries of $Z$ are highly correlated. This construction provides a neat way of incorporating a fixed matrix $X$ into the design of a RE matrix, and also has the advantage that storing $Z$ (as the factorization pair $(X\Phi^\top,\Phi)$) takes only $O(md)$ words of space, compared to storing a standard design matrix which typically requires $O(nd)$ words of space. In the next section, we will discuss few applications of this result to sparse linear regression problems.

We start with the definition of stable rank (denoted by $\sr()$) of a matrix $X$.   
$$\sr(X) = \norm{X}_\HS^2 / \norm{X}^2.$$
Stable rank cannot exceed the usual rank. The stable rank is a more robust notion than the usual rank because it is largely unaffected by tiny singular values.  In  Appendix~\ref{app:comp1} we provide a detailed comparison between these stable rank and restricted eigenvalue. Unsurprisingly, the picture that emerges is that stable rank is in fact a {\em less restrictive} condition.\!\footnote{In that a RE bound implies a non-trivial stable rank bound, whereas the other direction does not always hold.}  Throughout this section, $C, C_1, c, c_1, \ldots$ denote positive constants which may depend on the subgaussian norm of the entries of the involved matrices. 

We will work with a slightly modified (and a more general) definition of restricted eigenvalue that we state here. 
\begin{definition} \label{def:REmod}
Let $M$ be an $n \times d$ matrix, and let $k < d, \ \alpha>0$. Define
\[ \RE(M,k,\alpha)= \inf \frac{\norm{M\z}}{\norm{\z_J}}, \]
where $\z_J$ is the coordinate projection of $\z$ to $\R^J$, and  the infimum is taken over all sets $J \subset [d], \ |J|=k$ and all $\z \in \R^m \setminus \{0\}$ satisfying
 \[ \norm{\z_{J^\compl}}_1 \le \alpha \norm{\z_J}_1. \]
\end{definition}
Note that $\alpha = 3$ in Definition~\ref{def:RE}. Also given $\RE(M,k,\alpha)$, we can get a lower bound on $\xi$ in Definition~\ref{def:RE} as $ \xi \geq \RE(M,k,3)^2/k$. Our primary result in this section establishes a lower bound on $\RE(X \Phi^\top \Phi,k,\alpha)$. The proof assumes a stable rank condition on $X$ that we define below. The randomness of $\Phi$ makes the required condition on $X$ coordinate independent, unlike the RE condition which is tied to a concrete coordinate structure in $\R^d$. 

\begin{theorem} \label{thm:RE}
Let $m,n,d \in \N, \ m \le n \le d$, and let $X$ be a fixed $n \times d$ matrix satisfying
\begin{equation*} \mathbf{Stable\;Rank\;Condition:}\;\;\; 2 \le m \le \sr(X)/2. \end{equation*}
Let $\Psi=(\Psi_{ij})$ be an $m \times d$ random matrix with independent entries such that $\E[\Psi_{ij}] = 0$, $\E[\Psi_{ij}^2] = 1$, and $\|\Psi_{ij}\|_{\psi_2}$ is bounded. Let $\Phi = \Psi/\sqrt{m}$.
Let $\beta \in (0,1)$. Then for any $k \in \N, \ \alpha>0$ such that
\[ 1 \le \alpha \sqrt{k} \le \sqrt{\frac{cm}{k \log d + \log (2/\beta)}} \]
the matrix $X \Phi^\top \Phi$ satisfies
\[ \RE(X \Phi^\top \Phi,k,\alpha)  \ge \frac{\norm{X}_\HS}{32 \sqrt{m}}  \]
with probability at least $1-\beta$.
\end{theorem}
\begin{remark}
An intuitive explanation why stable rank is the correct notion here is as follows. Firstly, if $\rank(X\Phi^\top\Phi) \leq \rank (X) \leq k$, then RE does not hold for any $X\Phi^\top\Phi$. And it should be the stable rank, because adding an infinitesimally small noise does not change anything. The fact that we have the condition $m \succeq  k^2$ and not $m \succeq  k$, as this observation would suggest, is due to the model we considered, namely to the multiplication by $\Phi^{\top} \Phi$.
\end{remark}
\begin{corollary} \label{cor:RE}
Let $X$ and $\Psi$ be matrices satisfying the conditions in Theorem~\ref{thm:RE} with 
\[ 1 \le 3 \sqrt{k} \le \sqrt{\frac{cm}{k \log d + \log (2/\beta)}}. \]
Let $\Phi = \Psi/\sqrt{m}$. Then the matrix $X\Phi^\top \Phi$ satisfies:
$$ \inf_{S \subset [d], |S| = k, \theta \in \mathbb{C}(S)} \frac{ \| X\Phi^\top \Phi \theta \|^2}{n} \geq \frac{\| X \|_\HS^2 \| \theta \|^2}{1024 \, n m k},$$
with probability at least $1-\beta$.
\end{corollary}

We start with a high-level description of the proof idea. The complete proof is presented in Section~\ref{sec:RE}.  Let $\e_j$ denote the standard basis vector with $j$th entry set to $1$. 

\smallskip
\noindent\textbf{Idea of the Proof of Theorem~\ref{thm:RE}.} We now explain the idea behind the proof of the above theorem. Take any $J \subset [d], \ |J|=k$ and any $\y \in S^{d-1}$ with $\supp(\y) \subseteq J$. We wish to show that with overwhelming probability,  any $\x \in \R^d$ with $\supp(\x) \subseteq J^\compl$ and $\norm{\x}_1 \le \alpha \norm{\y}_1 \le \alpha \sqrt{k}$ satisfies 
\[ \norm{X \Psi^\top \Psi (\y+\x)} \ge r \]
 for some $r>0$. If the probability estimate is strong enough, we would be able to run an $\varepsilon$-net argument over all such $\y$ and take the union bound over all $J$ showing that $\RE(X\Psi^\top \Psi,k, \alpha) \ge r/2$. The condition above requires checking infinitely many $\x$. To make the problem tractable, let us introduce an orthogonal projection $Q: \R^n \to \R^n$ which we discuss more about later. Assume that $Q X \Psi^\top \Psi \y \neq 0$, and let $\u$ be the unit vector in the direction of $Q X \Psi^\top \Psi \y \neq 0$. Then
\begin{align*}
\norm{X \Psi^\top \Psi (\y+\x)} 
&\ge \norm{QX \Psi^\top \Psi (\y+\x)}
\ge \u^\top QX \Psi^\top \Psi (\y+\x) \\
&= \norm{QX \Psi^\top \Psi \y} + \u^\top QX \Psi^\top \Psi \x
\end{align*}
The quantity above is affine in $\x$, so it is minimized at one of the extreme points of the set $\{\x \in \R^d: \ \supp(\x) \subseteq J^\compl, \ \norm{\x}_1 \le \alpha \sqrt{k} \}$, i.e., at a vector $\pm\alpha \sqrt{k} \e_j, j \in J^\compl$. This observation allows us to pass from an infinite set of $\x$'s to a finite set. 

Next, we have to establish the concentration bounds on $\norm{QX \Psi^\top \Psi \y}$ and $\u^\top QX \Psi^\top \Psi \e_j$. Notice that $\Psi \y$ and $\Psi \e_j$ are independent centered (mean $0$) subgaussian vectors with the unit variance of the coordinates. If these vectors were independent of the random matrix $\Psi^\top$ as well, we would have  used the Hanson-Wright inequality to derive the necessary concentration. However, this is obviously not the case. At this moment, the projection $Q$ comes to the rescue. The idea is to carefully construct the projection to take care of the dependencies.


\subsection{Proof of Theorem~\ref{thm:RE}} \label{sec:RE}
In this section, we present the complete proof of Theorem~\ref{thm:RE}. In Section~\ref{sec: preliminaries}, we use the Hanson-Wright theorem and its corollaries to get  probabilistic estimates for norms of certain matrix products. In Section~\ref{sec: fixed vector}, we prove Theorem~\ref{thm:RE} for a fixed vector of a special form.  We finish the proof in Section~\ref{sec: proof}.

\subsubsection{Hanson-Wright Preliminaries} \label{sec: preliminaries}
We start by establishing probability estimates for the spectral and Frobenius norms for certain matrix products. The results in this section form the basic building blocks that are used throughout the proof. An important tool used here is the Hanson-Wright inequality and its several consequences. Hanson-Wright inequality establishes the concentration of a quadratic form of independent centered subgaussian random variables. An original (slightly weaker) version of this inequality was first proved in~\citep{hanson1971bound}. 

\begin{theorem}[Hanson-Wright Inequality \citep{RVHanson-Wright}] \label{thm: HW}
Let $\x = (x_1,\ldots,x_n) \in \R^n$ be a random vector with independent components $x_i$ which satisfy $\E[x_i] = 0$ and $\|x_i\|_{\psi_2}$ is bounded. Let $A$ be an $n \times n$ matrix. Then, for every $t \ge 0$,
\[ \Prob{ \left |\x^\top A \x - \E[\x^\top A \x] \right | > t} \le 2 \exp \Big( - c \min \Big( \frac{t^2}{ \norm{A}_\HS^2}, \frac{t}{ \norm{A}} \Big) \Big ). \]
\end{theorem}

Besides the theorem itself, we need several corollaries.
\begin{corollary}[Spectral Norm of the Product]	\label{cor: product norm}
Let $B$ be a fixed $n \times d$ matrix, and let $G=(G_{ij})$ be an $m \times d$ random matrix with independent entries that satisfy: $\E[G_{ij}] = 0$, $\E[G_{ij}^2] = 1$, and $\|G_{ij}\|_{\psi_2}$ is bounded. Then for any $s,t \ge 1$,
\[ \Prob{ \norm{BG^\top} > C (s \norm{B}_\HS+ t \sqrt{m} \norm{B}) } \le 2 \exp(- s^2 \sr(B) -  t^2 m) \]
and
\[ \Prob{ \norm{BG^\top} < \frac{1}{2} \norm{B}_\HS } \le 2 \exp(- c \sr(B)). \]
\end{corollary}
Corollary \ref{cor: product norm} can be found in \citep{RVHanson-Wright}. Assuming that $m \le \sr(B)$, we can rewrite the above inequalities as
\begin{equation}  \label{eq: product norm concentration}
\Prob{ \frac{1}{2} \norm{B}_\HS  <  \norm{BG^\top} < C \norm{B}_\HS } \ge 1- 2 \exp(- c \sr(B)).
\end{equation}
Applying this corollary in the case $m=1$, we obtain a small ball probability estimate for the image of a subgaussian vector. The small ball probability bounds the probability $\| B \g \|$ is small for a fixed matrix $B$ and a subgaussian vector $\g$. 
\begin{corollary}[Concentration for the Norm of a Vector] \label{cor: small ball}
Let $B$ be a fixed $n \times d$ matrix, and let $\g=(g_1,\dots,g_d) \in \R^d$ be a random vector with independent entries that satisfy $\E[g_j] = 0$, $\E[g_j^2] = 1$, and $\|g_j\|_{\psi_2}$ is bounded. Then
\[ \Prob{ \frac{1}{2} \norm{B}_\HS  <  \norm{Bg} < C \norm{B}_\HS } \ge 1- 2 \exp(- c \sr(B)). \]
\end{corollary}

Using this inequality, we can easily derive a small ball probability estimate for the Frobenius norm.
\begin{corollary}[Frobenius Norm of the Product] \label{cor: product HS norm}
Let $B$ be a fixed $n \times d$ matrix, and let $G=(G_{ij})$ be an $m \times d$ random matrix with independent entries that satisfy: $\E[G_{ij}] = 0$, $\E[G_{ij}^2] = 1$, and $\|G_{ij}\|_{\psi_2}$ is bounded. Then
\[ \Prob{ \frac{1}{2} \sqrt{m} \norm{B}_\HS  <  \norm{BG^\top}_{\HS} < C \sqrt{m} \norm{B}_\HS } \ge 1- 2 \exp(- c \sr(B)).\]
\end{corollary}
\begin{proof}
Denote the rows of $G$ by $\gamma_1 \etc \gamma_m$. Then,
\[ \norm{BG^\top}_\HS= \left( \sum_{j=1}^m \norm{B \gamma_j}^2 \right)^{1/2}. \]
The right-hand side can be interpreted as the Euclidean norm of the image of the vector $\tilde{\gamma} \in \R^{dm}$ obtained by concatenation of the vectors $\gamma_1 \etc \gamma_m$ under the $nm \times dm$ block-diagonal matrix $\tilde{B}= \diag(B \etc B)$. The result follows from the Corollary \ref{cor: small ball}, since $\norm{\tilde{B}}_\HS^2= m \norm{B}_\HS^2$ implying $\norm{\tilde{B}}_\HS= \sqrt{m} \norm{B}_\HS$.
\end{proof}

We will need a similar estimate for the Frobenius norm of the triple product of the form $GHG^\top$, where $H$ is a positive semidefinite matrix. Let $\tr()$ denote the trace of a matrix. 
\begin{corollary}[Frobenius norm of the Triple Product] \label{cor: thiple product HS norm}
Let $H$ be a fixed $d \times d$ symmetric positive semidefinite matrix, and let $G=(G_{ij})$ be an $m \times d$ random matrix with independent entries that satisfy: $\E[G_{ij}] = 0$, $\E[G_{ij}^2] = 1$, and $\|G_{ij}\|_{\psi_2}$ is bounded.   If $m \le \tr (H)/\norm{H}$, then
\[ \Prob{ \norm{G H G^\top}_\HS \ge C  \sqrt{m}  \cdot \tr(H)  }  \le 4 \exp \left( -c \frac{\tr(H)}{\norm{H}} \right). \]
  \end{corollary}
\begin{proof}
Let $H^{1/2}$ be the square root of the matrix $H$. Since $\tr(H)=\norm{H^{1/2}}_{\HS}^2$, the assumption of the corollary reads $m \le \sr(H^{1/2})$.
By Corollary \ref{cor: small ball},
\[
  \Prob{  \norm{H^{1/2}G^\top} \ge C \norm{H^{1/2}}_\HS } \le 2 \exp(- c \sr(H^{1/2})).
\]
Similarly, Corollary \ref{cor: product HS norm} implies
 \[
 \Pr[{ \norm{H^{1/2}G^\top}_{\HS} \ge C \sqrt{m} \norm{H^{1/2}}_\HS }] \le 2 \exp(- c \sr(H^{1/2})).
 \]
 As $\norm{G H G^\top}_\HS \le \norm{H^{1/2}G^\top}_{\HS} \cdot \norm{H^{1/2}G^\top}$, we have 
\begin{align*}
 \Prob{ \norm{G H G^\top}_\HS \ge C  \sqrt{m}  \cdot \norm{H^{1/2}}_{\HS}^2  } 
 &\le 4 \exp(- c \sr(H^{1/2}))  =  4 \exp \left( -c \frac{\tr(H)}{\norm{H}} \right),
\end{align*}
which completes the proof.
\end{proof}

\subsubsection{Bounds for a Fixed Vector} \label{sec: fixed vector}
In this section, our goal will be to investigate a special case of Theorem~\ref{thm:RE}. In particular, we investigate the RE condition in Definition~\ref{def:REmod} when restricted to vectors of the kind $\z = \e_j + \x$ for a fixed $j$ where $j \notin \supp(\x)$ (Proposition~\ref{prop: single direction}). The proof is based on two technical lemmas that use careful conditioning arguments along with the probabilistic inequalities established in the previous section. We use $\conv()$ and $\text{span}()$ to denote the convex hull and span of a set of vectors. We use $\text{Ker}()$ to denote the kernel of a matrix.

The following lemma bounds the small ball probability of $BG^\top \g$, for a fixed matrix $B$, random matrix $G$, and a random vector $\g$.
\begin{lemma} \label{lem: small ball product}
Let $B$ be a fixed $n \times d$ matrix, let $G=(G_{ij})$ be an $m \times d$ random matrix with independent entries and let $\g=(g_1,\dots,g_m) \in \R^m$ be a random vector with independent entries that satisfy: $\E[G_{ij}]=\E[g_j] = 0$, $\E[G_{ij}^2] = \E[g_j^2] = 1$, and $\|G_{ij}\|_{\psi_2}, \|g_j\|_{\psi_2}$ are bounded. Then
\[\Prob{ \norm{BG^\top \g} < \frac{1}{4} \sqrt{m} \norm{B}_\HS } \le 8 \Big( \exp \big(- c \sr(B) \big) + \exp(-cm) \Big). \]
\end{lemma}
\begin{proof}
Conditioning on $G$ and applying Corollary \ref{cor: small ball}, we obtain
\[ \Prob{ \norm{BG^\top \g} \le \frac{1}{2} \norm{BG^\top}_{\HS} \mid G} \le 2 \exp(-c \sr(BG^\top)). \]
Define the events $\Omega_\HS$ and $\Omega_{op}$ as in Corollary \ref{cor: thiple product HS norm}:
\begin{align*}
\Omega_\HS &= \left \{G: \ \norm{BG^\top}_\HS \ge \frac{1}{2} \sqrt{m} \norm{B}_\HS \right \} \\
 \Omega_{op} &= \left \{G: \ \norm{BG^\top} \le C (  \norm{B}_\HS + \sqrt{m} \norm{B} ) \right \}
\end{align*}
Let $\Omega_\HS^\compl$ and $\Omega_{op}^\compl$ denote the complement of these events respectively. Then by Corollaries \ref{cor: product HS norm} and \ref{cor: product norm},
\begin{align*}
& \Prob{ \norm{BG^\top \g} \le \frac{1}{4} \sqrt{m} \norm{B}_{\HS} } \\
&\le\Prob{ \norm{BG^\top \g} \le \frac{1}{2} \norm{BG^\top}_{\HS} \mid G \in \Omega_\HS \cap \Omega_{op}} +\Prob{\Omega_\HS^\compl} +\Prob{\Omega_{op}^\compl} \\
&\le 2 \exp \left(- c \frac{m \norm{B}_\HS^2}{ \norm{B}_\HS^2 + m \norm{B}^2} \right)+4 \exp(- c \sr(B)) \\
&\le 8 \Big( \exp \big(- c \sr(B) \big) + \exp(-cm) \Big).
 \end{align*}
\end{proof}

The following lemma provides a large deviation bound for a certain product form.
\begin{lemma}  \label{lem: product large deviation}
Let $B$ be a fixed $n \times d$ matrix, let $G=(G_{ij})$ be an $m \times d$ random matrix with independent entries and let $\g_1=(g_{1_1},\dots,g_{1_m}) \in \R^m$ and $\g_2=(g_{2_1},\dots,g_{2_m}) \in \R^m$ be random vectors with independent entries that satisfy: $\E[G_{ij}]=\E[g_{l_j}] = 0$, $\E[G_{ij}^2] = \E[g_{l_j}^2] = 1$, and $\|G_{ij}\|_{\psi_2},\|g_{l_j}\|_{\psi_2}$ are all bounded for $l \in \{1,2\}$. Assume that $m \le \sr(B)$. Then  for any $t \in \left[ 0, m \norm{B}_\HS^2 \right]$,
\[ \Prob{ |\g_1^\top G B^\top B G^\top \g_2| \ge t } \le 10 \exp \left(-c\frac{t^2}{ m \norm{B}_\HS^4} \right). \]
\end{lemma}
\begin{proof}
Define the vector $\g \in \R^{2m}$ and the $2m \times 2m$ matrix $\Gamma$ by
\[
\g= \begin{pmatrix}
\g_1 \\ \g_2
\end{pmatrix},
\qquad \Gamma=  \begin{pmatrix}
0 & G B^\top B  G^\top \\ G B^\top B  G^\top & 0
\end{pmatrix}.
\]
Condition on $G$. By Theorem \ref{thm: HW}, for any $t \ge 0$,
\[ \Prob{|\g^\top \Gamma \g| > t} \le 2 \exp \Big[ - c \min \Big( \frac{t^2}{ \norm{\Gamma}_\HS^2}, \frac{t}{ \norm{\Gamma}} \Big) \Big]. \]
Note that $\norm{\Gamma}=\norm{G B^\top B G^\top }= \norm{B G^\top}^2$. Let $\Omega_F$ and $\Omega_{op}$ be the events defined by
 \begin{align*}
\Omega_F &=\{G: \ \norm{G B^\top B G^\top }_\HS \le C \left(m \norm{B^\top B}_\HS + \sqrt{m}  \cdot \tr(B^\top B) \right) \} \\
\Omega_{op} &=\{G: \ \frac{1}{4} \norm{B}_\HS^2 \le  \norm{G B^\top B G^\top } \le C \norm{B}_\HS^2  \}
\end{align*}
Again, let $\Omega_\HS^\compl$ and $\Omega_{op}^\compl$ denote the complement events.  For any $G \in \Omega_\HS$,
\[\norm{\Gamma}_\HS^2 \le C  m  \cdot \tr(B^\top B)^2 = C'm \norm{B}_\HS^4.\]
Notice that 
\[ \frac{\tr(B^\top B)}{\norm{B^\top B}}= \sr(B).\]
 
Finally, combining this with Corollary \ref{cor: thiple product HS norm}, and \eqref{eq: product norm concentration}, we obtain
 \begin{align*}
 &\Prob{ |g_1^\top G B^\top B G^\top g_2| \ge t} \\
 &\le 2 \exp \Big[ - c \min \Big( \frac{t^2}{ m \norm{B}_\HS^4}, \frac{t}{ \norm{B}_\HS^2} \Big) \Big]
 + \Prob{\Omega_\HS^\compl} + \Prob{\Omega_{op}^\compl} \\
 &\le 4 \exp \left(-c\frac{t^2}{ m \norm{B}_\HS^4} \right) +  6 \exp(- c \sr(B)) 
 \end{align*}
 for any $t \in \left[ 0, m \norm{B}_\HS^2 \right]$.
 Since $m \le \sr(B)$, the first term  in the right-hand side dominates the second one, and the proof is complete.
 \end{proof}

Using Lemmas~\ref{lem: small ball product} and~\ref{lem: product large deviation}, we are ready to prove the following proposition. The main idea here is to introduce an orthogonal projection matrix which lets us decouple various dependencies that appear across various quantities.
\begin{proposition} \label{prop: single direction}
Let $R$ be a fixed $n \times d$ matrix, and let $G=(G_{i,j})$ be an $m \times d$ random matrix with independent entries that satisfy: $\E[G_{ij}] = 0$, $\E[G_{ij}^2] = 1$, and $\|G_{ij}\|_{\psi_2}$ is bounded. Assume that
\[ 2 \le m \le \sr(R)/2. \]
Then for any $s \ge 1$,
\begin{align*}
\Prob{ \exists \x \in s \cdot \conv(\pm \e_2 \etc \pm \e_d), \; \norm{RG^\top G (\e_1+\x)} \le \frac{1}{8} \sqrt{m} \norm{R}_\HS } &\le 2d \exp \left( -c \frac{m}{s^2} \right).
\end{align*}
\end{proposition}
\begin{proof}
Let $P_1$ be the orthogonal projection in $\R^n$ with $\text{Ker}(P_1)= \text{span}(R\e_1)$, where $\text{span}()$ denote the span.
Assume that $P_1 RG^\top G \e_1 \neq 0$ and set
\[ \u=\frac{P_1 RG^\top G \e_1}{\norm{P_1 RG^\top G \e_1}}. \]
Then
\begin{align} \label{eq: sd4}
\norm{RG^\top G (\e_1+\x)} &\ge \norm{P_1 RG^\top G (\e_1+\x)} \ge \norm{P_1 RG^\top G \e_1} - \u^\top P_1 RG^\top G \x. 
\end{align}
The minimal value of this expression over $\x \in s \cdot \conv(\pm \e_2 \etc \pm \e_d)$ is attained at the extreme points of this set. Consider $\x=s\e_2$ since all other extreme points are treated the same way. Since $\sr(R)>4$ and by the interlacing, we have
 \[ \norm{P_1R}_\HS^2 \ge \norm{R}_\HS^2 - \norm{R}^2 \ge \norm{R}_\HS^2/2 \]
 and so, $\sr(P_1R) \ge (1/2) \sr(R)$ (as $\| P_1 R \| = \| R\|$).

Denote by $\g_1$ and $\g_2$ the first and the second columns of $G$.  We have introduced $P_1$ to ensure that that the matrix $P_1RG^\top$ is independent of $\g_1$.  This allows us to replace the vector $\g_1$ by its copy independent of $G$. Hence, by Lemma~\ref{lem: small ball product},
\begin{align}  \label{eq: sd3}
\Prob{ \norm{P_1 RG^\top G \e_1} < \frac{1}{4} \sqrt{m} \norm{R}_\HS } &= \Prob{ \norm{P_1 RG^\top \g_1} < \frac{1}{4} \sqrt{m} \norm{R}_\HS }  \\
&\le 8 \Big( \exp \big(- c \sr(R) \big) + \exp(-cm) \Big) \le 2 \exp(-c' m), \notag
\end{align}
where we used that $m \le \sr(R)$.

The estimate of the inner product is a little more complicated. Let $P_2$ be the orthogonal projection with $\text{Ker}(P_2)=\text{span}(R\e_1, P_1R\e_2)$. Then we can write
\begin{align*}
P_1R G^\top G \e_1 &=P_2 R G^\top \g_1 + P_1R \e_2 \g_2^\top \g_1 \\
P_1R G^\top G \e_2 &=P_2 R G^\top \g_2 +  P_1R \e_2 \g_2^\top \g_2
\intertext{  and therefore,}
(P_1R G^\top G \e_1)^\top  P_1R G^\top G \e_2 & = (P_2 R G^\top \g_1)^\top P_2 R G^\top \g_2 + (P_1R \e_2 \g_2^\top \g_1)^\top P_1R \e_2 \g_2^\top \g_2.
\end{align*}

Note that $P_2 R G^\top$ is independent of $\g_1$ and $\g_2$. Similarly to~\eqref{eq: sd5}, we have
 \[ \norm{P_2R}_\HS^2 \ge \norm{R}_\HS^2 - 2\norm{R}^2 \ge \norm{R}_\HS^2/2
 \]
and so, $\sr(P_2 R) \ge (1/2) \sr(R) \ge m$. This allows us to use Lemma \ref{lem: product large deviation} to estimate
\begin{equation}\label{eq: sd1}
\Prob{ |\g_1^\top G (P_2R)^\top P_2R G^\top \g_2| \ge t } \le 8 \exp \left(-c\frac{t^2}{ m \norm{P_2R}_\HS^4} \right)
\end{equation}
for any $t \in [0, m \norm{P_2R}_\HS^2]$.

The estimate for the last term is straightforward as $P_1R \e_2$ is deterministic. Since
\[ \forall s \ge 0 \qquad \Prob{|\g_2^\top \g_1| > C s} \le 2 \exp \left(-c \frac{s^2}{m} \right) +\exp(-m), \]
and 
\[ \Prob{|\g_2^\top \g_2| > C m} \le \exp(-m), \]
we obtain
\begin{align}\label{eq: sd2}
&\Prob{|(P_1R \e_2 \g_2^\top \g_1 )^\top  P_1R \e_2 \g_2^\top \g_2| \ge s m \norm{P_1R\e_2}^2 } 
\le 2 \exp \left(-c \frac{s^2}{m} \right) + \exp(-m)  \notag
\intertext{or}
&\Prob{|(P_1R \e_2 \g_2^\top \g_1 )^\top  P_1R \e_2 \g_2^\top \g_2| \ge t } 
\le 2 \exp \left(-c \frac{t^2}{m^3 \norm{P_1R \e_2}^4} \right) + \exp(-m)   
\end{align}
for all $t \ge 0$.
Combining \eqref{eq: sd1} and \eqref{eq: sd2}, we conclude that
\begin{align*}
\Prob{| (P_1R G^\top G \e_1)^\top  P_1R G^\top G \e_2 | >t}  &\le  2 \exp \left(-c \frac{t^2}{m \norm{R}_\HS^4} \right) + 2 \exp \left(-c \frac{t^2}{m^3 \norm{P_1R \e_2}^4} \right) + \exp(-c m) \\
&\le  4 \exp \left(-c \frac{t^2}{m \norm{R}_\HS^4} \right) + \exp(-c m)
\end{align*}
for any $t \in [0, m \norm{P_2R}_\HS^2]$. Here we used the inequality
\[ m \norm{P_1R \e_2}^2 \le m \norm{R}^2 \le  \norm{R}_\HS^2, \]
where the last one follows from the assumption $m \le \sr(R)$. Taking into account the result from~\eqref{eq: sd3}, we see that
\[ \Prob{| \u^\top P_1 RG^\top G \e_2 | > \tau} \le  2 \exp \left(-c \frac{\tau^2}{ \norm{R}_\HS^2} \right) + \exp(-c m), \]
for all $\tau \in [0, \frac{1}{8}\sqrt{m} \norm{R}_\HS]$. After taking the union bound, we show that
\begin{align} \label{eq: sd5}
\Prob{\exists j \ge 2, \ | \u^\top P_1 RG^\top G \e_j | > \tau} \le  2d \left( \exp \left(-c \frac{\tau^2}{ \norm{R}_\HS^2} \right) + \exp(-c m) \right). 
\end{align}
Recall~\eqref{eq: sd4}. Setting $\tau= \frac{1}{8s}\sqrt{m} \norm{R}_\HS$ with $s \ge 1$, and using together~\eqref{eq: sd3} and~\eqref{eq: sd5}, we conclude that
\begin{align*}
\Prob{ \exists \x \in s \cdot \conv(\pm \e_2 \etc \pm \e_d), \; \norm{RG^\top G (\e_1+\x)} \le \frac{1}{8} \sqrt{m} \norm{R}_\HS } \le 2d \exp \left( -c \frac{m}{s^2} \right),
 \end{align*}
as the second term in the right-hand side gets absorbed in the first one. The proof of the proposition is complete.
\end{proof}

\subsubsection{Finishing the Proof of Theorem~\ref{thm:RE}: Net Argument} \label{sec: proof}
The next theorem is the main technical step in proving Theorem \ref{thm:RE}. Invoking this theorem with appropriate parameters (that we explain later in this section) gives the proof of Theorem~\ref{thm:RE}. The proof of the following theorem is based on generating an orthogonal matrix to reduce the general case to the special case discussed in Proposition~\ref{prop: single direction},  and then employing an $\varepsilon$-net argument. 
\begin{theorem} \label{thm: global} Let $X$ be a fixed $n \times d$ matrix satisfying,
\[ 2 \le m \le \sr(X)/2. \]
 Let $\Psi=(\Psi_{ij})$ be an $m \times d$ random matrix with independent entries such that $\E[\Psi_{ij}] = 0$, $\E[\Psi_{ij}^2] = 1$, and $\|\Psi_{ij}\|_{\psi_2}$ is bounded. Let $\beta \in (0,1)$, and let $k \in \N$. Then for any $s$ such that
\[ 1 \le s \le \sqrt{\frac{cm}{k \log d + \log (2/\beta)}}, \]
\begin{multline*}
\Pr  [\exists I \subset [d] \mbox{ with } |I|=k, \exists \y \in \mathbb{S}^{d-1} \mbox{ with } \supp (\y) \subseteq I,  \exists \x \in s \cdot \conv(\pm \e_i, \ i \notin I), \\ 
\norm{X \Psi^\top \Psi (\y+\x)} \le \frac{1}{32} \sqrt{m} \norm{X}_\HS ] \le \beta.
\end{multline*}
\end{theorem}
Note that the condition $s \ge 1$ in the formulation of the theorem implicitly sets a lower bound on $\beta$ and an upper bound on $k$.
\begin{proof}
Fix the set $I$ with $|I|=k$. For instance, consider $I=[k] \subset [d]$. Fix also a point $\y \in \mathbb{S}^{k-1}$. Define the subspace $E \subset \R^d$ as
\[E= \Span (\y, \e_j, \ j>k).\]
Note that the vectors $\y$ and $\e_j, \ j>k$ form an orthonormal basis of $E$.  Let $P_E: \R^d \to E$ be matrix of the orthogonal projection onto $E$ with respect to this basis and the standard basis in $\R^d$. Then $P_E^\top$ is the matrix of the embedding of $E$ into $\R^d$.

Let $Q: \R^n \to \R^n$ be the  orthogonal projection with $\text{Ker}(Q)=X E^\perp$, where $E^\perp$ represents the orthogonal complement of $E$. Then for any $\z \in E$,
\begin{equation}\label{eq: global1}
\norm{X \Psi^\top \Psi \z} \ge \norm{Q X \Psi^\top \Psi \z}.
\end{equation}
We can represent the restriction of the linear operator $Q X \Psi^\top \Psi$ to $E$ as the following composition of linear operators:
\[ E \overset{P_E^\top}{\to} \R^d \overset{\Psi}{\to} \R^m \overset{\Psi^\top}{\to} \R^d \overset{P_E}{\to} E \overset{P_E^\top}{\to} \R^d \overset{X}{\to} \R^n \overset{Q}{\to} \R^n. \]
Since $\norm{\y}=1$ and $\supp (\y) \subseteq [k]$, the $m \times (d-k+1)$ matrix $G=\Psi P_E^\top$ in the basis $\{\y, \e_j, \ j>k\}$ has centered subgaussian entries of unit variance. Denote $R=QX P_E^\top$. Then by the interlacing
\[ \norm{X}_\HS^2 \ge \norm{R}_\HS^2 \ge \norm{X}_\HS^2 - 2k \norm{X}^2 \ge \frac{1}{2} \norm{X}_\HS^2, \]
since by the assumptions on $k$ and $X$, $k \le m/8  \le \sr(X)/16$. This implies
 \[ \sr(R) \ge \frac{1}{2} \sr(X) \ge m. \]
 
Applying Proposition \ref{prop: single direction} to the matrices $G, R$, with $\y$ playing the role of $\e_1$, and taking into account \eqref{eq: global1}, we obtain
\begin{align*}
\Prob{ \exists \x \in s \cdot \conv(\pm \e_j \ j>k), \; \norm{X \Psi^\top \Psi (\y+\x)} \le \frac{1}{16} \sqrt{m} \norm{X}_\HS }  \le 2d \exp \left( -c \frac{m}{s^2} \right)
\end{align*}
for any $s \ge 1$.

In the rest of the proof, we employ the net argument. Since $\Psi$ is a subgaussian random matrix,
\begin{align*}
\norm{X \Psi^\top \Psi} &\le \norm{X \Psi^\top} \cdot \norm{\Psi} \le C' ( \norm{X}_{\HS}+ \sqrt{m} \norm{X}) \cdot C'' (\sqrt{d}+ \sqrt{m}) \\
&\le C \sqrt{d} \norm{X}_{\HS}
\end{align*}
with probability at least $1-\exp(-m)$, where we used Corollary \ref{cor: product norm}. Let $\varepsilon>0$ be a number to be chosen later, and (by Proposition~\ref{prop:epsnet}) let $\NN \subset \mathbb{S}^{k-1}$ be an $\varepsilon$-net of cardinality 
\[ |\NN| \le \left( \frac{3}{\varepsilon} \right)^k. \]
Assume that for any $\y \in \NN$, and for any $\x \in s \cdot \conv(\pm \e_j \ j>k)$,
\[ \norm{X \Psi^\top \Psi (\y+\x)} \ge \frac{1}{16} \sqrt{m} \norm{X}_\HS. \]
Assume also that $\norm{X \Psi^\top \Psi} \le C \sqrt{d} \norm{X}_{\HS}$. Let $\z \in \mathbb{S}^{k-1}$, and chose $\y \in \NN$ such that $\norm{\z-\y}<\varepsilon$. Then setting $\varepsilon= c \sqrt{m/d}$ for an appropriately small constant $c>0$, we obtain
\begin{align*}
\norm{X \Psi^\top \Psi (\z+\x)} &\ge \norm{X \Psi^\top \Psi (\y+\x)} -\norm{X \Psi^\top \Psi} \cdot \norm{\z-\y} \ge \frac{1}{32} \sqrt{m} \norm{X}_\HS.
\end{align*}
Thus,
\begin{align*}
& \Pr \left [\exists \y \in \mathbb{S}^{k-1},  \  \exists \x \in s \cdot \conv(\pm \e_i, \ i>k), \; \norm{X \Psi^\top \Psi (\y+\x)} \le \frac{1}{32} \sqrt{m} \norm{\Psi}_\HS \right ] \\ 
& \le |\NN| \cdot 2d \exp \left( -c \frac{m}{s^2} \right) + \exp(-m) \\
&\le 2 \exp \left( -c \frac{m}{s^2} + k \log \left( \frac{C \sqrt{d}}{\sqrt{m}} \right) \right).
\end{align*}
It remains to take the union bound over all possible supports of $\y$. It yields,
\begin{align*}
&\Pr [\exists I \subset [d] \mbox{ with } |I|=k,  \exists \y \in \mathbb{S}^{d-1} \mbox{ with } \supp (\y) \subseteq I,  \exists \x \in s \cdot \conv(\pm \e_i, \ i \notin I), \\
& \quad \quad \norm{X \Psi^\top \Psi (\y+\x)} \le \frac{1}{32} \sqrt{m} \norm{\Psi}_\HS  ] \\
&\le \binom{d}{k} \cdot 2 \exp \left( -c \frac{m}{s^2} + k \log \left( \frac{C \sqrt{d}}{\sqrt{m}} \right) \right) \\
&\le2 \exp \left( -c \frac{m}{s^2} + \frac{k}{2} \log \left( \frac{C d^2}{mk} \right) \right).
\end{align*}
The last quantity is smaller than $\beta$ provided that\footnote{Here we ignored smaller order terms assuming $d^2 \gg mk$. If this does not hold, one can obtain a slightly better estimate.}
\[ 1 \le s \le \sqrt{\frac{cm}{k \log d + \log (2/\beta)}}. \]
This completes the proof of the theorem.
\end{proof}

We now have all the ingredients to complete the proof of Theorem \ref{thm:RE}.
\begin{proof}[\textbf{Proof of Theorem \ref{thm:RE}}]
Assume that the complement of the event described in Theorem~\ref{thm: global} occurs. Namely, assume that
\begin{align*}
&\forall I \subset [d] \mbox{ with } |I|=k, \forall \y \in \mathbb{S}^{d-1} \mbox{ with }  \supp (\y) \subseteq I,  \forall \x \in s \cdot \conv(\pm \e_i, \, i \notin I ) \\
& \norm{X \Psi^\top \Psi (\y+\x)} \ge \frac{1}{32} \sqrt{m} \norm{X}_\HS.  
\end{align*}
If $s$ satisfies the condition of this theorem, then the event above occurs with probability at least $1-\beta$. Pick any $I \subset [d] \ |I|=k$ and any $\z \in \R^d \setminus \{0 \}$ with 
\[ \norm{\z_{I^\compl}}_1 \le \alpha \norm{\z_I}_1. \]
Without loss of generality, we may assume that $\y = \z_I \in \mathbb{S}^{d-1}$. Then, $\norm{\y}_1 \le \sqrt{k}$, and so $\norm{\z_{I^\compl}}_1 \le \alpha \sqrt{k}$. Theorem \ref{thm:RE} now follows from Theorem \ref{thm: global} applied with $s=\alpha \sqrt{k}$ and by plugging $\Phi = \Psi/\sqrt{m}$.
\end{proof}

\section{Applications to Sparse Linear Regression}
We now discuss some applications of our RE bound to the setting of sparse linear regression. We consider two different problems: \renewcommand{\labelenumi}{(\alph{enumi})}\begin{inparaenum} \item first one involves a standard regression setting with $Z=X\Phi^\top \Phi$ acting as a random  matrix, and the goal is to estimate the sparse $\theta^\star$ from a noisy linear model of observations \item second one is a variant of sparse linear regression, where the algorithm has access not to the individual covariates, but rather only to a randomly projected version of them, and the goal is to minimize (in-sample) prediction error.\end{inparaenum}

\subsection{Application 1:  Bounding the $\ell_2$-error with Random Design $Z=X \Phi^\top \Phi$} \label{sec:app1}
Consider the linear model $\y = Z \theta^\star + \w$, where $\w$ is an i.i.d.\ subgaussian noise. The following proposition establishes a $\ell_2$-error bound on estimating $\theta^\star$, using the standard Lasso analysis framework from Theorem~\ref{thm:lassoanalysis}. This result shows that $\ell_1$-relaxations succeed in estimating $\theta^\star$ even for certain dependent design matrices, partially justifying an observation commonly noticed in practice of Lasso succeeding even when the entries of the design matrix has dependencies. We work with a Lasso formulation defined on the pair $(Z,\y)$;

\begin{equation} \label{eqn:lassomod}
\theta^\comp \in  \mbox{argmin}_{\theta \in \R^d}\, \frac{1}{n} \| \y - Z \theta \|^2 + \lambda \| \theta \|_1 =  \mbox{argmin}_{\theta \in \R^d}\, \frac{1}{n}  (y_i - \langle \Phi \x_i, \Phi \theta \rangle)^2 +  \lambda \| \theta \|_1.
\end{equation}
The following proposition states the convergence bound of $\theta^\comp$ to $\theta^\star$ under this linear model. The probability in this case is over both the noise realization $\w$ and the randomness in $\Phi$. For brevity, we will say that the event which holds with probability at least $1-O(m^{-K})$ occurs with a {\em large} probability. 
\begin{proposition} \label{prop:app1}
Let $X$ be a deterministic matrix and $\Phi$ be a random matrix satisfying the conditions of Theorem~\ref{thm:RE}. Consider the linear model $\y = X \Phi^\top \Phi \theta^\star + \w$ where the entries of the noise vector $\w=(w_1,\dots,w_n)$ are independent centered subgaussians with $\| w_i \|_{\psi_2} \leq \sigma$. Let $K > 0$ be any constant, and let $d m^{-K} \leq \beta < 1$. Then $\theta^\comp \in \mbox{argmin}_{\theta \in \R_2^d}\, \| \y - X\Phi^\top\Phi \theta \|^2/n + \lambda\| \theta_1\|$ with $\lambda = \Theta(\sigma \norm{X}_F/n\sqrt{m} )$, satisfies with probability at least $1-\beta$: 
$$\| \theta^\comp -\theta^\star\| = O \left( \frac{\sigma \sqrt{m} k^{3/2}}{\| X \|_F}  \right ).$$
\end{proposition}
\begin{proof}
We use the framework of Theorem~\ref{thm:lassoanalysis} to bound $\|\theta^\comp - \theta^\star \|$.  The  matrix of interest is now $X \Phi^\top \Phi$. Our first aim will be to bound $\norm{\Phi^\top \Phi X^\top \w}_\infty$ that is used to set $\lambda$ in Theorem~\ref{thm:lassoanalysis}. Take any $\u \in S^{d-1}$. Then conditioning on $\Phi$, with a large probability,
\[ |\u^\top \Phi^\top \Phi X^\top \w| \le C \sigma \norm{\u^\top \Phi^\top \Phi X^\top}_2. \]
Applying the Hanson-Wright inequality, we can show that with a large probability with respect to $\Phi$,
\[ \norm{\u^\top \Phi^\top \Phi X^\top}_2 \le C \frac{\norm{X}_F}{\sqrt{m}}. \]
The estimate for $\norm{\Phi^\top \Phi X^\top \w}_\infty$ follows by combining two  previous inequalities and using the union bound for $\u=\e_j, \ j \in [d]$ as before. We get that probability at least $1-O(dm^{-K})$ ($\geq 1- O(\beta)$),
$$\norm{\Phi^\top \Phi X^\top \w}_\infty = O \left ( \sigma \frac{\norm{X}_F}{\sqrt{m}} \right).$$
Plugging this bound into Theorem~\ref{thm:lassoanalysis} along with the RE bound from Corollary~\ref{cor:RE} gives the claimed result.
\end{proof}

\subsection{Application 2: Sparse Linear Regression with Compressed Features} \label{sec:compsparse}
In this section, we use the results from Section~\ref{sec:REfromSR} on a sparse linear regression in a model where the regression algorithm only gets access to a compressed representation of the $\x_i$'s in the form of $\Phi \x_i$'s, and not to $\x_i$'s.\!\footnote{Throughout this section, we will assume that $\Phi$ is known to the algorithm.} As discussed in Section~\ref{sec:intro}, these compressed representations of the $\x_i$'s are easier to communicate in a distributed data setting and also reduces the storage requirements as we work with the compressed data. Consider the linear model $\y = X \theta^\star + \w$, where $X$ is some deterministic matrix and  $\w$ is subgaussian noise. Note that this is a fixed design setting (unlike the application in Section~\ref{sec:app1}).

Since the linear model is  $\y = X \theta^\star + \w$, and we only assume a rather weak stable rank assumption on $X$, getting an error bound on $\theta^\star$ is ruled out because as shown by~\citep{raskutti2011minimax} a condition closely related to restricted eigenvalue is needed for any parameter recovery method.%
\!\footnote{One simple illustration of why a stable rank condition on $X$ is not enough for parameter recovery, is that $\sr(X) \geq m$ (for some $m$) does not rule $X\theta^\star = 0$, which means $\y = \w$ implying $\y$ provides no information about $\theta^\star$, making any recovery of $\theta^\star$ impossible.} 
Therefore, in this section, we measure the performance in terms of minimizing {\em mean-squared (in-sample) prediction error}. Given $(\Phi\x_1,y_1),\dots,(\Phi\x_n,y_n)$, the goal is to output $\theta \in \R^d$ that has a relatively low prediction error $\| X \theta - X \theta^\star\|^2/n$. In a matrix-vector form, $(\Phi\x_1,y_1),\dots,(\Phi\x_n,y_n)$ can be represented as $(X\Phi^\top,\y)$. Now in a traditional sparse linear regression setting (with access to $(\x_1,y_1),\dots,(\x_n,y_n)$) this minimization can be performed without any assumptions on the design matrix $X$ (with faster convergence bounds possible under the RE assumption)~\citep{bickel2009simultaneous}. However, under the compressed setup, {\em a priori} it is unclear whether this seemingly simpler problem can even be solved consistently. 

A first idea given only $\Phi\x_i$'s will be to: \renewcommand{\labelenumi}{(\alph{enumi})}\begin{inparaenum} \item for all $i$, construct $\hat{\x}_i$, an approximation to $\x_i$ from $\Phi\x_i$, \item use the Lasso formulation on the resulting $(\hat{\x}_i,y_i)$'s\end{inparaenum}. This idea, however, is problematic because good reconstruction of $\x_i$'s from $\Phi \x_i$'s will require some strong (sparsity-like) assumptions on the structure of the $\x_i$'s, that is generally untrue. Another idea will be to construct an estimator $\hat{\vartheta} \in \R^m$ in the projected space say by minimizing the squared loss between $\y$ and $X \Phi^\top \vartheta$ (over $\vartheta \in \R^m$). This minimization would correspond to a different linear model: $\y = X \Phi^\top \vartheta^\star + \hat{\w}$. Since the true linear model is $\y = X \theta^\star + \w$, this would mean the new noise vector $\hat{\w} = (X \theta^\star - X \Phi^\top \vartheta^\star) +\w$ is no longer i.i.d.\ subgaussian. For bounding the prediction error (which in this case means bounding the norm of the difference between $X \Phi^\top \hat{\vartheta}$ and $X \theta^\star$) this could be problematic.  Additionally, given $\hat{\vartheta}$, lifting it to $\R^d$ is problematic as $\hat{\vartheta}$ may not be close to a projection of a sparse vector in $\R^d$. We overcome these hurdles by working with a Lasso formulation defined on the pair $(Z=X\Phi^\top\Phi,\y)$ as in~\eqref{eqn:lassomod}. Again define: $\theta^\comp \in \mbox{argmin}_{\theta \in \R^d}\,  \| \y - X \Phi^\top \Phi \theta \|^2/n + \lambda \| \theta \|_1$. Our basic idea is to establish a bound on $\| Z \theta^\comp - Z\theta^\star\|$, with the error vector $(\theta^\comp - \theta^\star)$ satisfying the cone set condition, and then using the RE bound on $Z$ to relate $\| Z \theta^\comp- Z\theta^\star\|$ and $\| X\theta^\comp - X\theta^\star\|$.

We start with the following inequality:
$$ \frac{1}{n} \| \y - Z \theta^\comp \|^2 + \lambda \| \theta^\comp \|_1 \leq \frac{1}{n} \| \y - Z \theta^\star \|^2 + \lambda \| \theta^\star \|_1.$$
Rearranging this gives
$$\frac{1}{n}(\| Z \theta^\comp \|^2  - \| Z \theta^\star \|^2) \leq  \frac{1}{n} \y^\top (Z \theta^\comp - Z \theta^\star) + \lambda (  \| \theta^\star \|_1 - \| \theta^\comp \|_1)$$
and plugging in $\tilde{\w} = \y - Z \theta^\star$ (i.e., $ \tilde{\w}=(X-Z) \theta^\star+\w$),
$$\frac{1}{n} \| Z \theta^\comp - Z \theta^\star \|^2 \leq \frac{1}{n} \tilde{\w}^\top(Z \theta^\comp - Z \theta^\star) + \lambda (  \| \theta^\star \|_1 - \| \theta^\comp \|_1).$$
Rearranging the terms, we get,
$$\frac{1}{n} \| Z \theta^\comp - Z \theta^\star \|^2 \leq \frac{1}{n} \langle Z^\top \tilde{\w},  \theta^\comp -  \theta^\star \rangle + \lambda (  \| \theta^\star \|_1 - \| \theta^\comp \|_1).$$
Adding and subtracting $\langle\E[Z^\top \tilde{\w}], \theta^\comp -  \theta^\star \rangle$ on the right-hand side gives,
$$\frac{1}{n} \| Z \theta^\comp - Z \theta^\star \|^2 \leq \frac{1}{n} \langle Z^\top \tilde{\w} - \E[Z^\top \tilde{\w}],  \theta^\comp -  \theta^\star \rangle + \langle  \E[Z^\top \tilde{\w}],  \theta^\comp -  \theta^\star \rangle + \lambda (  \| \theta^\star \|_1 - \| \theta^\comp \|_1).$$
By applying H\"older's inequality,
\begin{equation} \label{eqn:holder}
\frac{1}{n} \| Z \theta^\comp - Z \theta^\star \|^2 \leq \frac{1}{n} \| Z^\top \tilde{\w}  - \E[Z^\top \tilde{\w}] \|_\infty \| \theta^\comp - \theta^\star\|_1 + \| \E[Z^\top \tilde{\w}] \| \| \theta^\star -  \theta^\comp \| + \lambda (  \| \theta^\star \|_1 - \| \theta^\comp \|_1).
\end{equation}

The following lemma establishes a bound on $\| Z^\top \tilde{\w}  - \E[Z^\top \tilde{\w}] \|_\infty$. For simplicity, we focus on Gaussian random matrices $\Phi$, the extension of the lemma to a more general class of subgaussian random matrices is possible, but omitted here. Also w.l.o.g.\ we assume that $\theta^\star \in S^{d-1}$. The proof involves careful analysis of the projections of $Z^\top (X-Z) \theta^\star$ onto $\theta^\star$ and a vector in its orthogonal direction.

\begin{lemma} \label{lem: noise bound}
Let $X$ be an $n \times d$ matrix. Let $\Psi$ be an $m \times d$ standard Gaussian matrix with independent entries, and let $\Phi=\Psi/\sqrt{m}$.  Let $Z = X \Phi^\top \Phi$. Let $\w=(w_1,\dots,w_n) \in \R^n$ be a vector with independent centered coordinates having $\norm{w_j}_{\psi_2} \le \sigma$. Let $\theta^\star \in S^{d-1}$.
\[ \tilde{\w}=(X-Z) \theta^\star+\w. \]
Assume that $\sr(X) \ge m$. Then
\[ \E[Z^\top \tilde{\w}]= - \frac{\norm{X}_F^2}{m} \theta^\star \cdot (1+\delta), \]
where $\delta$ depends on $\theta^\star$ and $|\delta| \le 1/m$. and for any $K>0$, with probability at least $1-O(d m^{-K})$,
\[ \norm{Z^\top \tilde{\w}- \E[Z^\top \tilde{\w}]}_{\infty} \le C \log m \left( \frac{\norm{X}_F^2}{m^{3/2}}+ \sigma \frac{\norm{X}_F}{\sqrt{m}} \right) \]
with a constant $C$ depending on $K$. 
\end{lemma}
\begin{proof}
 Assume first that $\theta^\star=\e_1$. We will remove this assumption later. Set
 \[ \z:=\Phi^\top \Phi X^\top (X-X \Phi^\top \Phi) \e_1. \]
 To estimate $\z$ we consider projections of $\z$ on $\e_1$ and on a vector $\v \in S^{d-1}$ orthogonal to $\e_1$ separately. For brevity, we will say that the event which holds with probability at least $1-O(m^{-K})$ occurs with a {\em large} probability.
 \begin{step}
We will show that
\begin{equation} \label{eq: expected}
\E[\e_1^\top \z]= - \frac{\norm{X}_F^2}{m}  \cdot (1+\delta),
\end{equation}
where $|\delta| \le 1/m$ and with a large probability,
\[ |\e_1^\top \z- \E[\e_1^\top \z]| \le C \log m  \frac{\norm{X}_F^2}{m^{3/2}}. \] 
\end{step}
We will start with estimating $\e_1^\top \z$. To this end, denote $\Psi=(\upsilon_1 \ G)$ and $X=(\f_1 \ Y)$ separating the first column in each matrix. Then
\begin{align*}
\e_1^\top \z & = \frac{1}{m}\upsilon_1^\top (\upsilon_1 \ G)  X^\top \f_1 \cdot \left(1-\frac{1}{m}\upsilon_1^\top \upsilon_1 \right) - \frac{1}{m^2}\upsilon_1^\top (\upsilon_1 \ G) X^\top Y G^\top \upsilon_1 \\
& =: A+B.
\end{align*}
Let us estimate $A$ first. By Gaussian concentration, with a large probability, for a constant (independent of the parameters) $C$,
\[ |1- \frac{1}{m} \upsilon_1^\top \upsilon_1| \le C \frac{1}{\sqrt{m}}. \]
Hence,
\[ \left| \frac{1}{m} (\upsilon_1^\top \upsilon_1) \f_1^\top \f_1 \left (\frac{1}{m}- \upsilon_1^\top \upsilon_1 \right ) \right| \le C \frac{1}{\sqrt{m}} \norm{\f_1}_2^2 \le C \frac{1}{m^{3/2}} \norm{X}_F^2, \]
 where we used $\sr(X) \ge m$ in the last inequality. Also, conditioning on $G$, we also have that with large probability
\[ \left|\frac{1}{m}\upsilon_1^\top  G  Y^\top \f_1 \left(1-\frac{1}{m}\upsilon_1^\top \upsilon_1 \right) \right| \le C m^{-3/2} \sqrt{\log m} \norm{GY^\top \f_1}_2. \]
By Corollary~\ref{cor: product HS norm},
\[ \Pr \left [\norm{GY^\top \f_1}_2 \ge C \sqrt{m} \norm{Y^\top \f_1}_2 \right ] \le \exp(-m), \]
so with a large probability,
\begin{align*}
\left|\frac{1}{m}\upsilon_1^\top  G  Y^\top \f_1 \cdot \left(1-\frac{1}{m}\upsilon_1^\top \upsilon_1 \right) \right| \le C m^{-1} \sqrt{\log m}  \norm{Y^\top \f_1}_2 \le C m^{-3/2} \sqrt{\log m}  \norm{X}_F^2.
\end{align*}
Summarizing, we proved that with a large probability,
\[ |A| \le C m^{-3/2} \sqrt{\log m}  \norm{X}_F^2. \]

We move to estimating $B$. Denote 
\[ B= -\frac{1}{m^2} (\upsilon_1^\top \upsilon_1) \f_1^\top Y G^\top \upsilon_1 -\frac{1}{m^2} \upsilon_1^\top G Y^\top Y G^\top \upsilon_1 =: B_1+B_2. \]
Then $\E[B_1]=0$ and
\begin{align*}
\E[B_2] &= -\frac{1}{m^2} \E[\upsilon_1^\top GY^\top YG^\top \upsilon_1] = -\frac{1}{m^2} \tr ( \E[G Y^\top Y G^\top]) \\
&= -\frac{1}{m} \tr ( Y^\top Y ) = -\frac{\norm{Y}_F^2}{m}.
\end{align*}
Note that by assumption on $\sr(X)$,
 \[ \norm{X}_F^2-\norm{Y}_F^2=\norm{\f_1}_2^2 \le \frac{1}{m}\norm{X}_F^2, \]
 which yields \eqref{eq: expected} with the required bound on $\delta$.

Now, let us bound the deviation of $B$ from its expectation. Arguing as above, we conclude that with a large probability,
\[ |B_1| \le C m^{-1/2} \sqrt{\log m} \norm{\f_1^\top Y}_2 \le C m^{-3/2} \sqrt{\log m} \norm{X}_F^2. \]
Also, conditioning on $G$, with a large probability,
\begin{equation} \label{eq: B_2 1}
\left| \upsilon_1^\top G Y^\top Y G^\top \upsilon_1 - \E [ \upsilon_1^\top G Y^\top Y G^\top \upsilon_1 \mid G]\right| \le C \sqrt{\log m} \norm{G Y^\top Y G^\top}_F.
\end{equation}
Using the Hanson-Wright inequality as in Corollary~\ref{cor: thiple product HS norm}, we conclude that with a large probability,
\begin{align} \label{eq: B_2 2}
\norm{G Y^\top Y G^\top}_F \le Cm \sqrt{\log m}\norm{Y^\top Y}_F+C \sqrt{m} \tr(Y^\top Y) \le C \sqrt{m} \sqrt{\log m}\norm{X}_F^2
\end{align}
since $m \norm{Y^\top Y}_F \le m \norm{Y} \cdot \norm{Y}_F \le \sqrt{m} \norm{X}_F^2 \le \sqrt{m} \norm{X}_F^2$. The measure concentration with respect to the Gaussian matrix $G$ yields that with a large probability
\begin{align} \label{eq: B_2 3} 
\left| \E [\upsilon_1^\top G Y^\top Y G^\top \upsilon_1 \mid G]- \E[\upsilon_1^\top G Y^\top Y G^\top \upsilon_1] \right|  \le C m \norm{Y^\top Y}_F \le C \sqrt{m} \norm{X}_F^2.
\end{align} 
Combining \eqref{eq: B_2 1}, \eqref{eq: B_2 2}, and \eqref{eq: B_2 3} shows that with a large probability,
\[ |B_2 - \E[B_2]| \le C m^{-3/2} \log m \norm{X}_F^2. \]
This together with the bounds on $A$ and $B_1$ obtained above completes the proof of Step 1.
 
\begin{step}
Let $\v \in S^{d-1}, \ \v^\top \e_1=0$. Then $\E[\v^\top \z]=0$ and with a large probability,
\[ |\v^\top \z| \le C m^{-3/2} \log m \norm{X}_F^2. \]
\end{step}
The equality $\E[\v^\top \z]=0$ follows from independence of $\Phi \e_1$ and $\Phi \v$ (recall that we assumed that the matrix $\Psi$ is Gaussian).
 To prove the concentration, we can use rotation invariance of the Gaussian distribution. More precisely, the matrix $\Phi$ is distributed like $\Phi U$, where $U$ is an orthogonal matrix such that $U \e_1 =\e_1$ and $U\v=\e_2$. Note that replacing $X$ by $X U^\top$ does not change the Hilbert-Schmidt norm. Using these observations, we can reduce the case of a general $\v \perp \e_1$ to $\v=\e_2$.
 
In the last case, we separate the first two columns of the matrices $\Psi$ and $X$ as we did in Step 1:
\[ \Psi=( \upsilon_1 \ \upsilon_2 \  \Lambda) \quad \text{and} \quad X=(\f_1 \ \f_2 \ P). \]
Then the inner product $\e_2^\top \z$ can be decomposed into a sum of $9$ terms containing different combinations of independent random variables $\upsilon_1, \upsilon_2$, and $\Lambda$. The absolute value of each of these terms does not exceed $C m^{-3/2} \log m \norm{X}_F^2$ with a large probability. These estimates closely follow the argument of Step 1, so we omit the details. Combining these nine estimates completes the proof of Step 2.
 
Let us summarize what we proved. We have shown that in the case $\theta^\star=\e_1$, 
\[ \E[\z]= - \frac{\norm{X}_F^2}{m}  \cdot (1+\delta) \e_1 = - \frac{\norm{X}_F^2}{m}  \cdot (1+\delta) \theta^\star \]
and for any $\u \in S^{d-1}$, 
\[ |\u^\top (\z - \E[\z])| \le C m^{-3/2} \log m \norm{X}_F^2. \]
The last inequality follows by decomposing $\u$ into its projection on $\e_1$ and the orthogonal component and applying Steps 1 and 2 respectively to these components.

Now, we can use the invariance of the Gaussian matrix under multiplication by an orthogonal one to remove the assumption that $\theta^\star=\e_1$ in two last inequalities. To derive the bound for $\norm{\z- \E[z]}_{\infty}$, we apply the last inequality with $\u=\e_j$ and take the union bound over $j \in [d]$. 
 
Finally, it remains to handle the term $\norm{\Phi^\top \Phi X^\top \w}_\infty$ which we do as in Proposition~\ref{prop:app1}. As before, take any $\u \in S^{d-1}$. Then conditioning on $\Phi$, with a large probability,
\[ |\u^\top \Phi^\top \Phi X^\top \w| \le C \sigma \norm{\u^\top \Phi^\top \Phi X^\top}_2. \]
Applying the Hanson-Wright inequality, we show that with a large probability with respect to $\Phi$,
\[ \norm{\u^\top \Phi^\top \Phi X^\top}_2 \le C \frac{\norm{X}_F}{\sqrt{m}}. \]
The estimate for $\norm{\Phi^\top \Phi X^\top \w}_\infty$ follows by combining two  previous inequalities and using the union bound for $\u=\e_j, \ j \in [d]$ as before.
 
This completes the proof of the lemma.
\end{proof}

\begin{remark} \label{rem:follow}
The same argument shows that for any $K>0$, with probability at least $1-O(d^{-K})$,
\[ \norm{\Phi^\top \Phi X^\top \tilde{\w}- \E[\Phi^\top \Phi X^\top \tilde{\w}]}_{\infty} \le C \log d \left( \frac{\norm{X}_F^2}{m^{3/2}}+ \sigma \frac{\norm{X}_F}{\sqrt{m}} \right) \]
with a constant $C$ depending on $K$.
\end{remark}

Applying Lemma~\ref{lem: noise bound} in~\eqref{eqn:holder}, gives that with probability at least $1-O(dm^{-K})$,
\begin{multline} \label{eqn:condition}
\frac{1}{n} \| Z \theta^\comp - Z \theta^\star \|^2  \leq \frac{C\log m }{n}  \left ( \frac{\norm{X}_F^2}{m^{3/2}}+ \sigma \frac{\norm{X}_F}{\sqrt{m}} \right) \| \theta^\comp - \theta^\star\|_1 \\ + \frac{2 \norm{X}_F^2}{n m} \| \theta^\comp - \theta^\star\|_1 + \lambda (  \| \theta^\star \|_1 - \| \theta^\comp \|_1).
\end{multline}
For the remainder of this proof, we condition on~\eqref{eqn:condition} holding true.
Let $S_\star = \supp(\theta^\star)$. We first argue that $\hat{\theta} := \theta^\comp - \theta^\star$ is such that $\hat{\theta} \in \mathbb{C}(S_\star)$. We start by observing that:
\begin{align} \label{eqn:thetahat}
\| \theta^\star \|_1 - \| \theta^\comp \|_1 & = \| \theta^\star \|_1 - \| \theta^\star + \hat{\theta} \|_1 = \| \theta^\star \|_1 - \| \theta^\star_{S_\star} + \hat{\theta}_{S_\star} \|_1 - \| \hat{\theta}_{S_\star^\compl} \|_1 \leq \| \hat{\theta}_{S_\star} \|_1 -  \| \hat{\theta}_{S_\star^\compl} \|_1.
\end{align}
Set
$$\lambda \geq  \frac{2 C\log m }{n}  \left ( \frac{\norm{X}_F^2}{m^{3/2}}+ \sigma \frac{\norm{X}_F}{\sqrt{m}} \right) +  \frac{4 \norm{X}_F^2}{n m}.$$
Using this value of $\lambda$, we can observe that,
$$\frac{1}{n} \| Z \theta^\comp - Z \theta^\star \|^2  \leq \frac{\lambda}{2} \| \hat{\theta} \|_1 + \lambda (  \| \theta^\star \|_1 - \| \theta^\comp \|_1).$$
From~\eqref{eqn:thetahat} and by noting $\| Z \theta^\comp - Z \theta^\star \|^2 > 0$,
$$0 \leq \frac{\lambda}{2} \| \hat{\theta} \|_1 + \lambda (\| \hat{\theta}_{S_\star} \|_1 -  \| \hat{\theta}_{S_\star^\compl} \|_1),$$
implying $\| \hat{\theta}_{S_\star^\compl} \|_1 \leq 3 \| \hat{\theta}_{S_\star} \|_1$, i.e., $\hat{\theta} \in \mathbb{C}(S_\star)$.
We can now simplify~\eqref{eqn:condition} as,
$$ \frac{1}{n} \| Z \hat{\theta} \|^2  = O \left ( \frac{ \| X \|_\HS^2 \log m }{n m^{3/2}} + \frac{\sigma  \| X \|_\HS \log m}{n \sqrt{m}} + \frac{\| X \|_\HS^2}{nm} \right ) \|   \hat{\theta} \|_1.$$
Now,
$$ \| \hat{\theta} \|_1 = \| \hat{\theta}_{S_\star} \|_1 +  \| \hat{\theta}_{S_\star^\compl} \|_1 \leq \hat{\theta}_{S_\star} \|_1 + 3 \| \hat{\theta}_{S_\star} \|_1 \leq 4 \sqrt{k} \| \hat{\theta}_{S_\star} \| \leq 4 \sqrt{k} \| \hat{\theta}\| \leq 8 \sqrt{k},$$
as $\| \hat{\theta}\| = \| \theta^\comp - \theta^\star \| \leq \| \theta^\comp \| + \| \theta^\star \|  \leq 2$.\!\footnote{Since $\theta^\star \in S^{d-1}$, it suffices to define $\theta^\comp \in \mbox{argmin}_{\theta \in B_2^d}\, \| \y - Z \theta \|^2/n + \lambda\| \theta_1\|$, implying that $\| \theta^\comp \| \leq 1$.}
Plugging this in the above inequality,
$$ \frac{1}{n} \| Z \hat{\theta} \|^2  = O \left ( \frac{ \| X \|_\HS^2 \sqrt{k}  \log m}{n m^{3/2}} + \frac{\sigma  \| X \|_\HS \sqrt{k} \log m}{n \sqrt{m}} + \frac{\| X \|_\HS^2 \sqrt{k}}{nm} \right ) \| \hat{\theta} \| = O \left ( \frac{\sigma  \| X \|_\HS \sqrt{k} \log m}{n \sqrt{m}} + \frac{\| X \|_\HS^2 \sqrt{k}}{nm}  \right ).$$

Now by our stable rank assumption on $X$, $\| X \hat{\theta} \| = O((\| X \|_\HS/\sqrt{m}) \| \hat{\theta} \|)$. Under the conditions of Corollary~\ref{cor:RE}, with probability at least $1-\beta$, $\| Z \hat{\theta} \|^2 = \Omega( (\| X \|_\HS^2/mk) \| \hat{\theta} \|^2)$. Putting these two together gives that, $\| X \hat{\theta} \|^2 =O( k \|  Z \hat{\theta} \|^2)$. 

Using this in the above bound on $\| Z \hat{\theta} \|^2 = \| Z(\theta^\comp - \theta^\star) \|^2$ gives that with probability at least $1-\beta$, under the conditioning on~\eqref{eqn:condition}:
$$\frac{1}{n} \| X\theta^\comp - X \theta^\star \|^2 =  O \left ( \frac{\sigma  \| X \|_\HS k^{3/2} \log m}{n \sqrt{m}} + \frac{\| X \|_\HS^2 k^{3/2}}{nm}  \right ).$$
Finally, we can remove the conditioning on~\eqref{eqn:condition}. To simplify the result, assume $\beta > d m^{-K}$.

\begin{proposition} \label{prop:main}\!\footnote{We assume that the algorithm has an estimate of $\| X \|_F$ (a good upper bound suffices). This is easy achievable in the distributed data setting described in Section~\ref{sec:intro} (Figure~\ref{fig:skconv}), as each device $i$ in addition to $(\Phi \x_i,y_i)$ can also communicate $\| \x_i\|$ to the server.}
Let $X$ be a deterministic matrix and $\Phi$ be a Gaussian random matrix satisfying the conditions of Theorem~\ref{thm:RE}.  Consider the linear model $\y = X \theta^\star + \w$ where the entries of the noise vector $\w=(w_1,\dots,w_n)$ are independent centered subgaussians with $\| w_i \|_{\psi_2} \leq \sigma$ and $\theta^\star \in S^{d-1}$. Let $K > 0$ be any constant, and let $d m^{-K} \leq \beta < 1$. Then $\theta^\comp \in \mbox{argmin}_{\theta \in B_2^d}\, \| \y - X\Phi^\top\Phi \theta \|^2/n + \lambda\| \theta_1\|$ with $\lambda = \Theta(\sigma \|X\|_\HS \log m/n\sqrt{m}+ \| X\|_\HS^2/nm)$, satisfies with probability at least $1-\beta$: 
$$\frac{1}{n} \| X\theta^\comp - X \theta^\star \|^2 =  O \left ( \frac{\sigma  \| X \|_\HS k^{3/2} \log m}{n \sqrt{m}} + \frac{\| X \|_\HS^2 k^{3/2}}{nm}  \right ).$$
\end{proposition}
\begin{remark}
For a small $\sigma$, the dominant term in the  error bound in Proposition~\ref{prop:main} is the $\| X \|_\HS^2 k^{3/2}/nm$ term. If we set, $m = \sr(X)/2$,  then $\| X \|_\HS^2 k^{3/2}/nm = 2\| X \|^2 k^{3/2}/n$, and therefore in this case we get a consistent prediction if $\| X \|  = o(\sqrt{n}/k^{3/4})$.  
\end{remark}


\begin{thebibliography}{28}
\providecommand{\natexlab}[1]{#1}
\providecommand{\url}[1]{\texttt{#1}}
\expandafter\ifx\csname urlstyle\endcsname\relax
  \providecommand{\doi}[1]{doi: #1}\else
  \providecommand{\doi}{doi: \begingroup \urlstyle{rm}\Url}\fi

\bibitem[Adamczak et~al.(2011)Adamczak, Litvak, Pajor, and
  Tomczak-Jaegermann]{adamczak2011restricted}
Radoslaw Adamczak, Alexander~E Litvak, Alain Pajor, and Nicole
  Tomczak-Jaegermann.
\newblock Restricted isometry property of matrices with independent columns and
  neighborly polytopes by random sampling.
\newblock \emph{Constructive Approximation}, 34\penalty0 (1):\penalty0 61--88,
  2011.

\bibitem[Bandeira et~al.(2016)Bandeira, Fickus, Mixon, and
  Moreira]{bandeira2016derandomizing}
Afonso~S Bandeira, Matthew Fickus, Dustin~G Mixon, and Joel Moreira.
\newblock Derandomizing restricted isometries via the legendre symbol.
\newblock \emph{Constructive Approximation}, 43\penalty0 (3):\penalty0
  409--424, 2016.

\bibitem[Bandeira et~al.(2017)Bandeira, Mixon, and
  Moreira]{bandeira2017conditional}
Afonso~S Bandeira, Dustin~G Mixon, and Joel Moreira.
\newblock A conditional construction of restricted isometries.
\newblock \emph{International Mathematics Research Notices}, 2017\penalty0
  (2):\penalty0 372--381, 2017.

\bibitem[Bickel et~al.(2009)Bickel, Ritov, and
  Tsybakov]{bickel2009simultaneous}
Peter~J Bickel, Ya'acov Ritov, and Alexandre~B Tsybakov.
\newblock Simultaneous analysis of lasso and dantzig selector.
\newblock \emph{The Annals of Statistics}, pages 1705--1732, 2009.

\bibitem[Bourgain et~al.(2011)Bourgain, Dilworth, Ford, Konyagin, Kutzarova,
  et~al.]{bourgain2011explicit}
Jean Bourgain, Stephen Dilworth, Kevin Ford, Sergei Konyagin, Denka Kutzarova,
  et~al.
\newblock Explicit constructions of rip matrices and related problems.
\newblock \emph{Duke Mathematical Journal}, 159\penalty0 (1):\penalty0
  145--185, 2011.

\bibitem[Candes et~al.(2007)Candes, Tao, et~al.]{candes2007dantzig}
Emmanuel Candes, Terence Tao, et~al.
\newblock The dantzig selector: Statistical estimation when p is much larger
  than n.
\newblock \emph{The Annals of Statistics}, 35\penalty0 (6):\penalty0
  2313--2351, 2007.

\bibitem[Candes and Tao(2005)]{candes2005decoding}
Emmanuel~J Candes and Terence Tao.
\newblock Decoding by linear programming.
\newblock \emph{IEEE transactions on information theory}, 51\penalty0
  (12):\penalty0 4203--4215, 2005.

\bibitem[Cheraghchi(2011)]{cheraghchi2011coding}
Mahdi Cheraghchi.
\newblock Coding-theoretic methods for sparse recovery.
\newblock In \emph{Communication, Control, and Computing (Allerton), 2011 49th
  Annual Allerton Conference on}, pages 909--916. IEEE, 2011.

\bibitem[De~Castro(2014)]{de2014optimal}
Yohann De~Castro.
\newblock Optimal designs for lasso and dantzig selector using expander codes.
\newblock \emph{IEEE Transactions on Information Theory}, 60\penalty0
  (11):\penalty0 7293--7299, 2014.

\bibitem[Dobriban and Fan(2016)]{dobriban2016regularity}
Edgar Dobriban and Jianqing Fan.
\newblock Regularity properties for sparse regression.
\newblock \emph{Communications in mathematics and statistics}, 4\penalty0
  (1):\penalty0 1--19, 2016.

\bibitem[Eldar and Kutyniok(2012)]{eldar2012compressed}
Yonina~C Eldar and Gitta Kutyniok.
\newblock \emph{Compressed sensing: theory and applications}.
\newblock Cambridge University Press, 2012.

\bibitem[Hanson and Wright(1971)]{hanson1971bound}
David~Lee Hanson and Farroll~Tim Wright.
\newblock A bound on tail probabilities for quadratic forms in independent
  random variables.
\newblock \emph{The Annals of Mathematical Statistics}, 42\penalty0
  (3):\penalty0 1079--1083, 1971.

\bibitem[Hastie et~al.(2015)Hastie, Tibshirani, and
  Wainwright]{hastie2015statistical}
Trevor Hastie, Robert Tibshirani, and Martin Wainwright.
\newblock \emph{Statistical learning with sparsity: the lasso and
  generalizations}.
\newblock CRC Press, 2015.

\bibitem[Lecu{\'e} and Mendelson(2017)]{lecue2017sparse}
Guillaume Lecu{\'e} and Shahar Mendelson.
\newblock Sparse recovery under weak moment assumptions.
\newblock \emph{Journal of the European Mathematical Society}, 19\penalty0
  (3):\penalty0 881--904, 2017.

\bibitem[Lee et~al.(2015)Lee, Sun, Liu, and Taylor]{lee2015communication}
Jason~D Lee, Yuekai Sun, Qiang Liu, and Jonathan~E Taylor.
\newblock Communication-efficient sparse regression: a one-shot approach.
\newblock \emph{arXiv preprint arXiv:1503.04337}, 2015.

\bibitem[Mendelson et~al.(2008)Mendelson, Pajor, and
  Tomczak-Jaegermann]{mendelson2008uniform}
Shahar Mendelson, Alain Pajor, and Nicole Tomczak-Jaegermann.
\newblock Uniform uncertainty principle for bernoulli and subgaussian
  ensembles.
\newblock \emph{Constructive Approximation}, 28\penalty0 (3):\penalty0
  277--289, 2008.

\bibitem[Negahban et~al.(2012)Negahban, Ravikumar, Wainwright, and
  Yu]{negahban2012unified}
Sahand~N Negahban, Pradeep Ravikumar, Martin~J Wainwright, and Bin Yu.
\newblock A unified framework for high-dimensional analysis of m-estimators
  with decomposable regularizers.
\newblock \emph{Statistical Science}, 27\penalty0 (4), 2012.

\bibitem[Oliveira(2016)]{oliveira2016lower}
Roberto~Imbuzeiro Oliveira.
\newblock The lower tail of random quadratic forms with applications to
  ordinary least squares.
\newblock \emph{Probability Theory and Related Fields}, 166\penalty0
  (3-4):\penalty0 1175--1194, 2016.

\bibitem[Raskutti et~al.(2010)Raskutti, Wainwright, and
  Yu]{raskutti2010restricted}
Garvesh Raskutti, Martin~J Wainwright, and Bin Yu.
\newblock Restricted eigenvalue properties for correlated gaussian designs.
\newblock \emph{JMLR}, 11:\penalty0 2241--2259, 2010.

\bibitem[Raskutti et~al.(2011)Raskutti, Wainwright, and
  Yu]{raskutti2011minimax}
Garvesh Raskutti, Martin~J Wainwright, and Bin Yu.
\newblock Minimax rates of estimation for high-dimensional linear regression
  over-balls.
\newblock \emph{Information Theory, IEEE Transactions on}, 57\penalty0
  (10):\penalty0 6976--6994, 2011.

\bibitem[Rudelson and Vershynin(2008)]{rudelson2008sparse}
Mark Rudelson and Roman Vershynin.
\newblock On sparse reconstruction from fourier and gaussian measurements.
\newblock \emph{Communications on Pure and Applied Mathematics}, 61\penalty0
  (8):\penalty0 1025--1045, 2008.

\bibitem[Rudelson and Vershynin(2013)]{RVHanson-Wright}
Mark Rudelson and Roman Vershynin.
\newblock Hanson-wright inequality and sub-gaussian concentration.
\newblock \emph{Electronic Communications in Probability}, 18, 2013.

\bibitem[Rudelson and Zhou(2013)]{rudelson2013reconstruction}
Mark Rudelson and Shuheng Zhou.
\newblock Reconstruction from anisotropic random measurements.
\newblock \emph{Information Theory, IEEE Transactions on}, 59\penalty0
  (6):\penalty0 3434--3447, 2013.

\bibitem[Sivakumar et~al.(2015)Sivakumar, Banerjee, and
  Ravikumar]{sivakumar2015beyond}
Vidyashankar Sivakumar, Arindam Banerjee, and Pradeep~K Ravikumar.
\newblock Beyond sub-gaussian measurements: High-dimensional structured
  estimation with sub-exponential designs.
\newblock In \emph{Advances in neural information processing systems}, pages
  2206--2214, 2015.

\bibitem[Tibshirani(1996)]{tibshirani1996regression}
Robert Tibshirani.
\newblock Regression shrinkage and selection via the lasso.
\newblock \emph{Journal of the Royal Statistical Society. Series B
  (Methodological)}, pages 267--288, 1996.

\bibitem[Tibshirani and Wasserman(2015)]{tibshiranisparsity}
Ryan Tibshirani and Larry Wasserman.
\newblock Sparsity and the lasso,
  \url{http://www.stat.cmu.edu/~larry/=sml/sparsity.pdf}, 2015.

\bibitem[Wainwright(2009)]{wainwright2009sharp}
Martin~J Wainwright.
\newblock Sharp thresholds for high-dimensional and noisy sparsity recovery
  using $\ell_1$-constrained quadratic programming (lasso).
\newblock \emph{Information Theory, IEEE Transactions on}, 55\penalty0 (5),
  2009.

\bibitem[Zhou et~al.(2009)Zhou, Lafferty, and Wasserman]{zhou2009compressed}
Shuheng Zhou, John Lafferty, and Larry Wasserman.
\newblock Compressed and privacy-sensitive sparse regression.
\newblock \emph{Information Theory, IEEE Transactions on}, 55\penalty0
  (2):\penalty0 846--866, 2009.

\end{thebibliography}

\appendix

\section{Additional Preliminaries} \label{app:addl}
\noindent\textbf{Background on Sparse Linear Regression.} If the linear model $\y=M\theta^\star+\w$, where $M \in \R^{n \times d}$ is high-dimensional in nature, meaning that the number of observations $n$ is substantially smaller than $d$, then it is easy to see that without further constraints on $\theta^\star$, the statistical model $\y = M \theta^\star + \w$ is not {\em identifiable}. This is because (even when $\w = 0$), there are many vectors $\theta^\star$ that are consistent with the observations $\y$ and $M$. This identifiability concern may be eliminated by imposing some type of sparsity assumption on the regression vector $\theta^\star$. Typically, $\theta^\star$ is $k$-sparse for $k \ll d$. 
Disregarding computational cost, the most direct approach to estimating a $k$-sparse $\theta$ in the linear regression model would be solving a quadratic optimization problem with an $\ell_0$-constraint:
\begin{align} \label{eqn:sparsereg}
\theta^{\text{sparse}} \in \mbox{argmin}_{\theta \in \Sigma_k}\, \frac{1}{n} \| \y - M \theta \|^2.
\end{align}

\paragraph{Lasso Regression.}  Since~\eqref{eqn:sparsereg} leads to a non-convex problem, a natural alternative is obtained by replacing the $\ell_0$-constraint with its tightest convex relaxation, the $\ell_1$-norm. This leads to the popular Lasso regression, defined as,
\begin{equation*} 
\mbox{Lasso Regression (penalized form):} \quad \theta^\Lasso \in \mbox{argmin}_{\theta \in \R^d}\, \frac{1}{n}\sum_{i=1}^n \| \y - M \theta \|^2.+  \lambda \| \theta \|_1,
\end{equation*}
for some choice $\lambda > 0$.

The consistency properties of Lasso are well-understood. Under a variety of mild assumptions on the instance, the Lasso estimator ($\theta^\Lasso$) is known to converge to the sparse $\theta^\star$ in the $\ell_2$-norm. Under stronger assumptions (such as mutual incoherence, minimum eigenvalue, and minimum signal condition) on the instance, it is also known that $\theta^\Lasso$ will have the same support as $\theta^\star$. We refer the reader to the recent book~\citep{hastie2015statistical} for a detailed survey of developments in this area.

\smallskip
\noindent\textbf{Background on $\varepsilon$-Nets.} Consider a subset $T$ of $\R^d$, and let $\varepsilon > 0$. A $\varepsilon$-net of $T$ is a subset $\NNN \subseteq T$ such that for every $\x \in T$, there exists a $\y \in \NNN$ such that $\| \x - \y \| \leq \varepsilon$.
\begin{proposition} [Volumetric Estimate] \label{prop:epsnet}
Let $T$ be a subset of $B_2^d$ and let $\varepsilon > 0$. Then there exists an $\varepsilon$-net $N$ of $T$ of cardinality at most $(1 + 2/\varepsilon)^d$. For any $\varepsilon \leq 1$, this can be simplified as $(1 + 2/\varepsilon)^d \leq (3/\varepsilon)^d$.
\end{proposition}

\smallskip
\noindent\textbf{Background on Subgaussian Random Variables.} Subgaussian random variables are a wide class of random variables, which contains in particular the standard normal, Bernoulli, and all bounded random variables.
\begin{definition} [Subgaussian Random Variable]\label{def:subgauss} 
We call a random variable $x \in \R$ subgaussian if there exists  a constant $C > 0$ if $\Pr[ |x| > t]  \leq 2 \exp(-t^2/C^2)$ for all $t > 0$. 
\end{definition}

\begin{definition}[Norm of a Subgaussian Random Variable] \label{def:subgaussnorm}
The $\psi_2$-norm of a subgaussian random variable $x \in \R$, denoted by $\| x \|_{\psi_2}$ is: $ \| x \|_{\psi_2} = \inf \left \{ t > 0 \,:\, \E[\exp(|x|^2/t^2)] \leq 2 \right \}$.
\end{definition}
Note that the $\psi_2$ condition on a scalar random variable $x$ is equivalent to the subgaussian tail decay of $x$.

\section{Comparison between Stable Rank and Restricted Eigenvalue Conditions} \label{app:comp1}
In this section, we investigate how stable rank relates to the restricted eigenvalue (RE) condition that is commonly used in the analysis of Lasso. The picture that emerges is the following: stable rank is a {\em less restrictive} condition to impose on a design matrix than RE. We show this by establishing that a RE bound on a matrix implies a non-trivial\footnote{A direct numerical extension is not possible as stable rank is invariant to matrix scaling, whereas RE is not.} stable rank for that matrix, whereas other direction does not always hold. 

We first look at the case, when we have a stable rank condition on $X$. The RE condition (and of course, RIP) governs the behavior of the matrix on \emph{all} coordinate subspaces of a small dimension. In this sense, a bound on the stable rank on $X$ is much more relaxed. We now provide a simple pedagogical example to illustrate this fact. We rely on the fact that if $X\e_j=0$ for even one $j \in d$, then no RE condition holds. Consider, for example the $d \times n$ matrix
\[ X=\begin{pmatrix}
\mathbb{I}_{2m} & 0 \\
0 & 0
\end{pmatrix},
\]
where $\mathbb{I}_{2m}$ is the identity $2m \times 2m$ matrix. Then, $\sr(X)=2m$, while the RE condition does not hold for $X$. This simple example illustrates that there exist families of matrices for which a stable rank condition (as required in Theorem~\ref{thm:RE}) holds, but a RE condition is not satisfied. 

To make the comparison in the other direction, we need an additional normalization of $X$, as $\sr(X)$ is invariant under scaling, and $\RE(X,k,\alpha)$ is degree $1$ homogenous (in that scaling each element in $X$ by a factor $c$ changes $\RE(X,k,\alpha)$ by $c$). Assume that $\RE(X,k,\alpha) \ge r$ and define
\[
 \norm{X}_{(k)} = \max_{\substack{J \subset [d] \\ |J|=k} } \norm{X_J} \le R.
\]
An upper bound on $\norm{X}_{(k)}$ is usually applied together with a lower bound on $\RE(X,k,\alpha) \ge r$ in derivation of the vector reconstruction conditions (see, e.g. \citep{rudelson2013reconstruction}). These assumptions yield that 
\[ \norm{X}_\HS = \left(\sum_{j=1}^d \norm{X\e_j}^2 \right)^{1/2} \ge r \sqrt{d}. \] 
Also, assume for simplicity that $d=kL$ and decompose $[d]=\bigcup_{l=1}^L J_l$, where $J_l \subset [d]$ are consecutive sets of $k$ coordinates. Let $\y \in \mathbb{S}^{d-1}$. Then
\begin{align*}
\norm{X \y}  \le \sum_{l=1}^L \norm{X_{J_l}} \cdot  \norm{\y_{J_l}} 
\le  \left( \sum_{l=1}^L \norm{X_{J_l}}^2 \right)^{1/2}  \left( \sum_{l=1}^L \norm{\y_{J_l}}^2 \right)^{1/2} \le R \sqrt{L} = R \sqrt{\frac{d}{k}}.
\end{align*}
Therefore,  $\norm{X}  \le  R \sqrt{\frac{d}{k}}$ and so 
\[
 \sr(X) \ge \left(\frac{r}{R} \right)^2 k.
\]
This shows that a RE bound on $X$ implies a non-trivial stable rank bound on $X$. 

Putting both these directions together implies that while a RE bound always translates into stable rank bound, the other direction does not always hold.

\end{document}